\documentclass[10pt,twocolumn,letterpaper]{article}

\usepackage{cvpr}
\usepackage{times}
\usepackage{epsfig}
\usepackage{graphicx}
\usepackage{amsmath}
\usepackage{amssymb}

\usepackage{cases}
\usepackage[linesnumbered,ruled,vlined,lined,boxed,commentsnumbered]{algorithm2e}
\usepackage{subfigure}
\usepackage{indentfirst}
\usepackage{multirow}
\usepackage{boldline}
\usepackage{booktabs}
\usepackage{multirow}

\usepackage[toc,page]{appendix}

\usepackage{enumitem}
\setitemize{noitemsep,topsep=0pt,parsep=0pt,partopsep=0pt}
\setenumerate{noitemsep,topsep=0pt,parsep=0pt,partopsep=0pt}

\usepackage[pagebackref=true,breaklinks=true,letterpaper=true,colorlinks,bookmarks=false]{hyperref}

\usepackage[usenames, dvipsnames]{color}

\cvprfinalcopy 

\def\cvprPaperID{1501} 

\ifcvprfinal\fi
\begin{document}

\title{Learning to Evaluate Image Captioning}

\author{
Yin Cui$^{1,2}$\hspace{15pt}Guandao Yang$^1$\hspace{15pt}Andreas Veit$^{1,2}$\hspace{15pt}Xun Huang$^{1,2}$\hspace{15pt}Serge Belongie$^{1,2}$\\
{$^1$Department of Computer Science, Cornell University\hspace{20pt}$^2$Cornell Tech}
}

\maketitle

\begin{abstract}
Evaluation metrics for image captioning face two challenges.
Firstly, commonly used metrics such as CIDEr, METEOR, ROUGE and BLEU often do not correlate well with human judgments.
Secondly, each metric has well known blind spots to pathological caption constructions, and rule-based metrics lack provisions to repair such blind spots once identified.
For example, the newly proposed SPICE correlates well with human judgments, but fails to capture the syntactic structure of a sentence.
To address these two challenges, we propose a novel learning based discriminative evaluation metric that is directly trained to distinguish between human and machine-generated captions.
In addition, we further propose a data augmentation scheme to explicitly incorporate pathological transformations as negative examples during training.
The proposed metric is evaluated with three kinds of robustness tests and its correlation with human judgments.
Extensive experiments show that the proposed data augmentation scheme not only makes our metric more robust toward several pathological transformations, but also improves its correlation with human judgments.
Our metric outperforms other metrics on both caption level human correlation in Flickr 8k and system level human correlation in COCO.
The proposed approach could be served as a learning based evaluation metric that is complementary to existing rule-based metrics.
\end{abstract}

\begin{figure}[t]
\begin{center}
\includegraphics[width=\columnwidth]{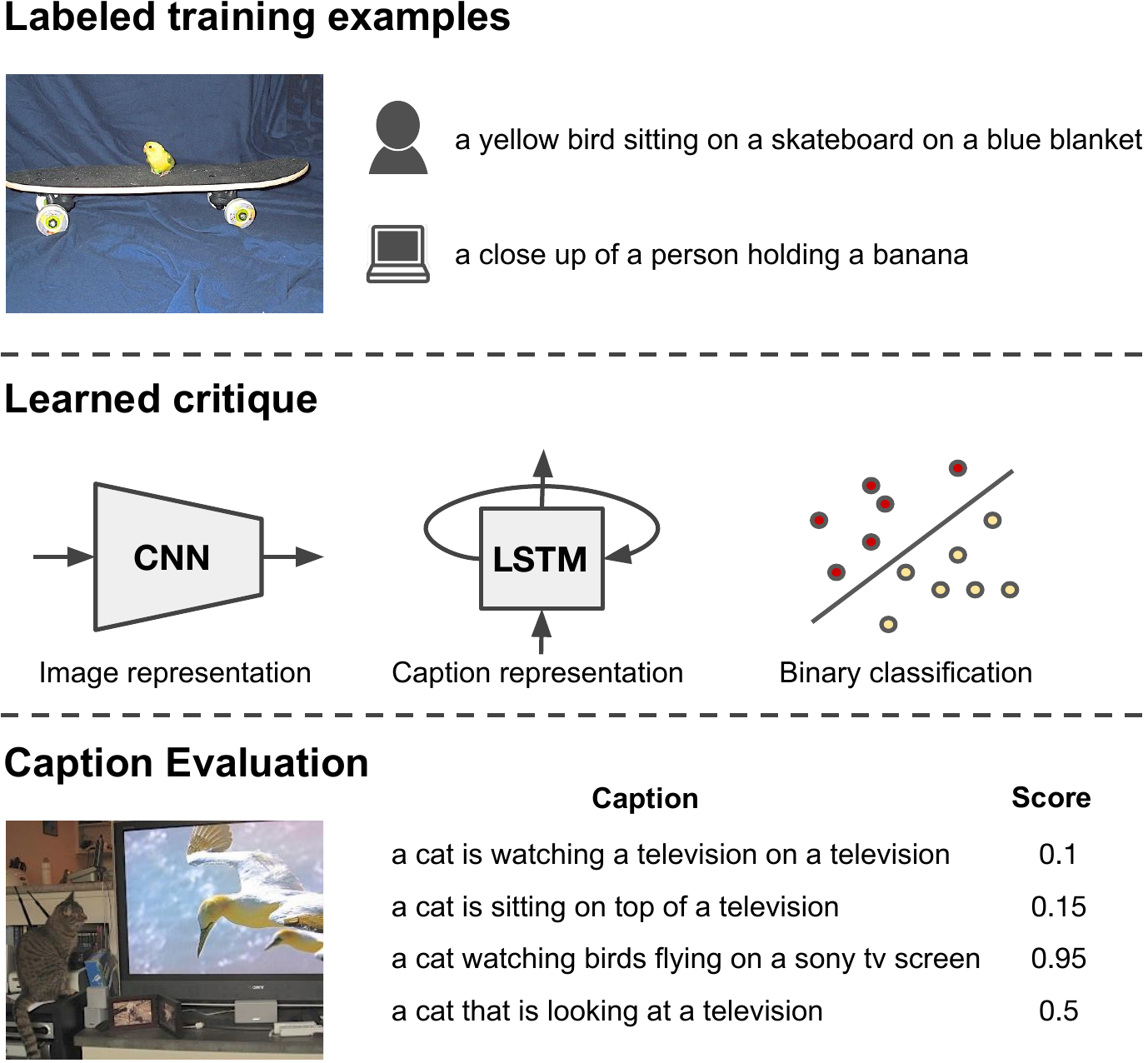}
\end{center}
\caption{
An overview of our proposed captioning evaluation metric. From a set of images and corresponding human written and machine generated captions, we train a model to discriminate between human and generated captions. The model comprises three major components: a CNN to compute image representations, an RNN with LSTM cells to encode the caption, and a binary classifier as the critique. After training, the learned critique can be used as a metric to evaluate the quality of candidate captions with respect to the context (\ie,~the image and reference human captions).
}
\label{fig:overview}
\end{figure}

\section{Introduction}
Learning to automatically generate captions to summarize the content of an image is considered as a crucial task in Computer Vision.
The evaluation of image captioning models is generally performed using metrics such as BLEU~\cite{papineni2002bleu}, METEOR~\cite{lavie2014meteor}, ROUGE~\cite{lin2004rouge} or CIDEr~\cite{vedantam2015cider}, all of which mainly measure the word overlap between generated and reference captions. The recently proposed SPICE~\cite{anderson2016spice} measures the similarity of scene graphs constructed from the candidate and reference sentence, and shows better correlation with human judgments.

These commonly used evaluation metrics face two challenges.
Firstly, many metrics fail to correlate well with human judgments.
Metrics based on measuring word overlap between candidate and reference captions find it difficult to capture semantic meaning of a sentence, therefore often lead to bad correlation with human judgments.
Secondly, each evaluation metric has its well-known blind spot, and rule-based metrics are often inflexible to be responsive to new pathological cases.
For example, SPICE is sensitive to the semantic meaning of a caption but tends to ignore its syntactic quality.
Liu~\etal~\cite{liu2017improved} shows that SPICE prefers to give high score to long sentences with repeating clauses.
It's not easy to let SPICE take such pathological cases into account.
Since it's difficult to completely avoid such blind spots, a good evaluation metric for image captioning should be flexible enough to adapt to pathological cases once identified, while correlating well with human judgments.

To address the aforementioned two challenges, we propose a metric that directly discriminates between human and machine generated captions while being able to flexibly adapt to pathological cases of our interests.
Since real human judgment is impractical to obtain at scale, our proposed learning based metric is trained to perform like a human critique, as illustrated in Fig.~\ref{fig:overview}.
We use a state-of-the-art CNN architecture to capture high-level image representations, and a RNN with LSTM cells to encode captions.
To design the learned critique, we follow insights from the COCO Captioning Challenge in 2015~\cite{cocochallenge, cococaptions}, in which a large-scale human judgment experiment was performed. In particular, our critique is a binary classifier that makes a Turing Test type judgment in which it differentiates between human-written and machine-generated captions.

In order to capture targeted pathological cases, we propose to incorporate these pathological sentences as negative training examples.
To systematically create such pathological sentences, we define several transformations to generate unnatural sentences that might get high scores in an evaluation metric.
Our proposed data augmentation (Sec.~\ref{sec:data_augmentation}) scheme uses these transformations to generate large number of negative examples, which guide our metric to explore a variety of possible sentence constructions that are rare to be found in real world data.
Further, we propose a systematic approach to measure the robustness of an evaluation metric to a given pathological transformation (Sec.~\ref{sec:capability_reliability}).
Extensive experiments (Sec.~\ref{sec:experiments}) verify the effectiveness and robustness of our proposed evaluation metric and demonstrate better correlation with human judgments on COCO and Flickr 8k, compared with commonly-used image captioning metrics.

Our key contributions can be summarized as follows:
\begin{itemize}
\item We propose a novel learned based captioning evaluation metric that directly captures human judgments while being flexible to targeted pathological cases.
\item We demonstrate key factors for how to successfully train a good captioning evaluation metric.
\item We conduct comprehensive studies that demonstrates the effectiveness of the proposed metric, in particular its correlation to human judgment and robustness toward pathological transformations.
\end{itemize}

\section{Related Work}



\textbf{Captioning evaluation}. 
Despite recent interests, image captioning is notoriously difficult to evaluate due to the inherent ambiguity. 
Human evaluation scores are reliable but costly to obtain. 
Thus, current image captioning models are usually evaluated with automatic metrics instead of human judgments.
Commonly used evaluation metrics BLEU~\cite{papineni2002bleu}, METEOR~\cite{lavie2014meteor}, ROUGE~\cite{lin2004rouge} and CIDEr~\cite{vedantam2015cider} are mostly based on n-gram overlap and tend to be insensitive to semantic information. 
Anderson \etal recently proposed the SPICE~\cite{anderson2016spice} that is based on scene graph similarity. 
Although SPICE obtains significantly higher correlation with human judgments, it encounters difficulties with repetitive sentences, as pointed out in~\cite{liu2017improved}.
It is worth noting that all above mentioned metrics rely solely on similarity between candidate and reference captions, without taking the image into consideration. Our proposed metric, on the other hand, takes image feature as input. While all the previous metrics are rule-based, our proposed metric learns to score candidate captions by training to distinguish positive and negative examples. Moreover, our proposed training scheme could flexibly take new pathological cases into account, yet traditional metrics find it hard to adapt.

\textbf{Adversarial training and evaluation}. 
Generative Adversarial Networks (GANs)~\cite{goodfellow2014generative} have been recently applied to generate image captions \cite{dai2017towards, liang2017recurrent, chen2017show, shetty2017speaking}. Although GANs could provide discriminators to tell apart human and machine generated captions, they differ from our works as our discriminator focuses on evaluation instead of generation.
All existing adversarial evaluation approaches define the generator performance to be inversely proportional to the classification performance of the discriminator, motivated by the intuition that a good generator should produce outputs that are hard for the discriminator to distinguish from real data.
The specific configurations differ among the approaches.
Im \etal~\cite{im2016generating} propose to train a pair of GANs and interchange their opponents during testing.
Iowe \etal~\cite{lowe2016towards} attempt to train a single discriminator on a large corpus of dialogue responses generated by different dialogue systems. 
Other approaches~\cite{kannan2017adversarial,bowman2016generating,li2017adversarial} train one discriminator  separately for each model.
Different from implicitly generated negative examples by a generator in these work, we incorporate explicitly defined pathological transformations to generate negative examples.
Moreover, none of the above literature has verified the effectiveness of their metrics by the correlation with human judgments.


\begin{figure*}[h!]
\centering
\subfigure{
\includegraphics[width=\linewidth]{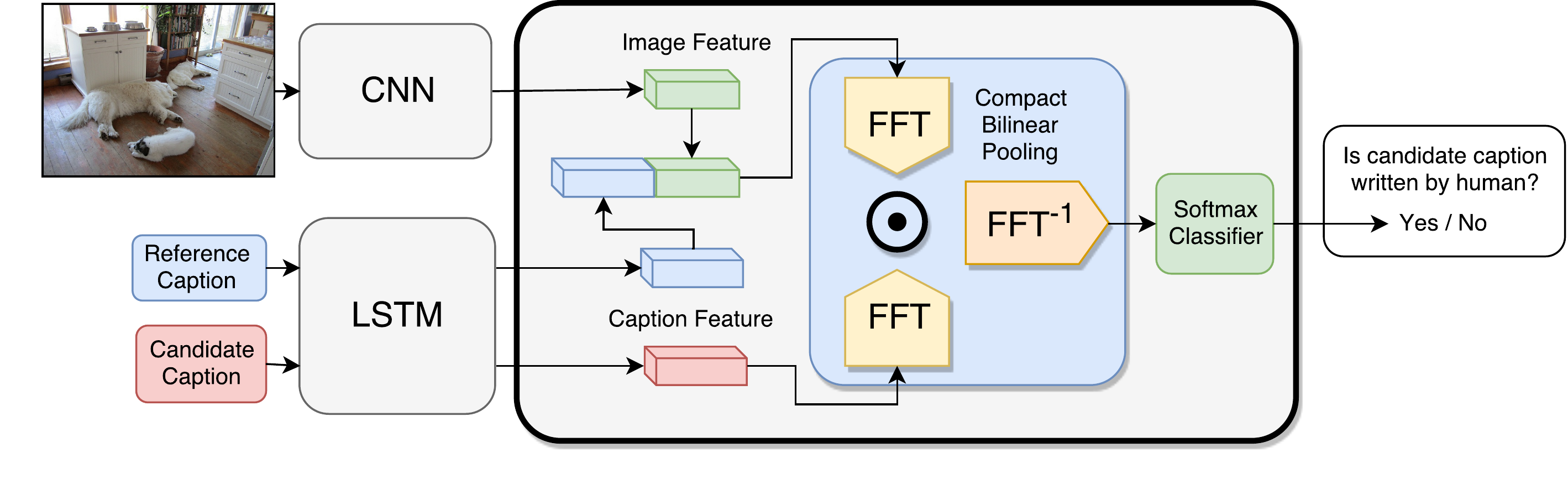}
}
\caption{
  The model architecture of the proposed learned critique with Compact Bilinear Pooling. We use a deep residual network and an LSTM to encode the reference image and human caption into context vector. The identical LSTM is applied to get the encoding of a candidate caption. The context feature and the feature extracted from the candidate caption are combined by compact bilinear pooling. 
  The classifier is supervised to perform a Turing Test by recognizing whether a candidate caption is human written or machine generated.
}
\label{fig:model_architecture}
\end{figure*}

\section{Discriminative Evaluation}
A caption is considered of high quality if it is judged well by humans.
In particular, the quality of a generated caption is measured by how successful it can fool a critique into believing it is written by human. 

\subsection{Evaluation Metric}
The proposed evaluation metric follows the general setup of a Turing Test. 
First, we train an automatic critique to distinguish generated captions from human-written ones. 
We then score candidate captions by how successful they are in fooling the critique.

Formally, given a critique parametrized by $\Theta$, a reference image $i$, and a generated caption $\hat{c}$, the score is defined as the probability for the caption of being human-written, as assigned by the critique:
\begin{equation}
    \text{score}_\Theta(\hat{c}, i) = P(\hat{c} \text{ is human written } |\ i, \Theta)
\end{equation}
The score is conditioned on the reference image, because the task of the evaluation is not simply to decide whether a given sentence is written by a human or machine generated, but also to evaluate whether it accurately captures the image content and focuses on the important aspects of the image.

More generally, the reference image represents the context in which the generated caption is evaluated. To provide further information about the relevance and salience of the image content, a reference caption can additionally be supplied to the context. 
Let $\mathcal{C}(i)$ denotes the context of image $i$, then reference caption $c$ could be included as part of context, \ie, $c \in \mathcal{C}(i)$.
The score with context becomes
\begin{equation}
    \text{score}_\Theta(\hat{c}, i) = P(\hat{c} \text{ is human written } |\ \mathcal{C}(i), \Theta)
\label{eqn:score}
\end{equation}

\subsection{Model Architecture}
\label{sec:model_architecture}
The proposed model can be generally described in two parts.
In the first part, the context information including the image and reference caption are encoded as feature vectors.
These two feature vectors are then concatenated as a single context vector.
In the second part, the candidate caption is encoded into a vector, in the same way as the reference caption.
We then fed it into a binary classifier, together with the context vector.
Fig.~\ref{fig:model_architecture} gives an overview of the model.

To encode the image $i$ as a feature vector $\mathbf{i}$, we use a ResNet~\cite{resnet} pre-trained on ImageNet~\cite{imagenet} with fixed weights.
The reference caption $c$ as well as the candidate caption $\hat{c}$ are encoded as feature vectors $\mathbf{c}$ and $\mathbf{\hat{c}}$ using an LSTM-based~\cite{lstm} sentence encoder. To form the input of the LSTM, each word is represented as a $d$-dimensional word embedding vector $\mathbf{x} \in \mathbb{R}^d$ which is initialized from GloVe~\cite{glove}. The LSTMs used to encode the two captions share the same weights.
The weights of the initial word embedding as well as of the LSTM are updated during training.

Once the encoded feature vectors are computed, they are combined into a single vector.
In our experiments, we use two different ways to combine these features; both methods provide comparable results.
The first method simply concatenates the vectors followed by a MLP:
\begin{equation}
    \mathbf{v} = \text{ReLU}(W\cdot\text{concat}([\mathbf{i}, \mathbf{c}, \mathbf{\hat{c}}]) + b)
\end{equation}
where $\text{ReLU}(x) = \max (x,0)$.
For the second method, we first concatenate the context information as $\text{concat}([\mathbf{i}, \mathbf{c}])$ and subsequently combine it with the candidate caption using Compact Bilinear Pooling (CBP) \cite{cbp}, which has been demonstrated in \cite{fukui2016multimodal} to be very effective in combining heterogeneous information of image and text.
CBP uses Count Sketch \cite{count_sketch, tensor_sketch} to approximate the outer product between two vectors in a lower dimensional space. This results in a feature vector $\mathbf{v}$ that captures $2$\textsuperscript{nd}\xspace order feature interactions compactly as represented by:
\begin{equation}
    \mathbf{v} = \Phi\big(\text{concat}([\mathbf{i}, \mathbf{c}])\big) \otimes \Phi\big(\mathbf{\hat{c}}\big)
\end{equation}
where $\Phi(\cdot)$ represents Count Sketch and $\otimes$ is the circular convolution.
In practice, circular convolution is usually calculated in frequency domain via Fast Fourier Transform (FFT) and its inverse (FFT$^{-1}$).

The feature combination is followed by a 2-way softmax classifier representing the class probabilities of being human written or machine generated. Finally, the classifier is trained using the cross-entropy loss function $\mathcal{H}(\cdot,\cdot)$:
\begin{equation}
    \mathcal{L} = \frac{1}{N}\sum_{n=1}^{N} \mathcal{H}(p_n,q_n)
\end{equation}
where $N$ is the number of training examples, $p$ is the output of the softmax classifier and $q$ is a one-hot vector indicating the ground truth of whether a candidate caption is indeed human written or machine generated.

By assigning a loss function that directly captures human judgment, the learned metric is capable of measuring the objective of the image captioning task. 
During inference, the probability from the softmax classifier of being the human written class is used to score candidate captions.

\begin{figure}[t]
\begin{center}
\includegraphics[width=\columnwidth]{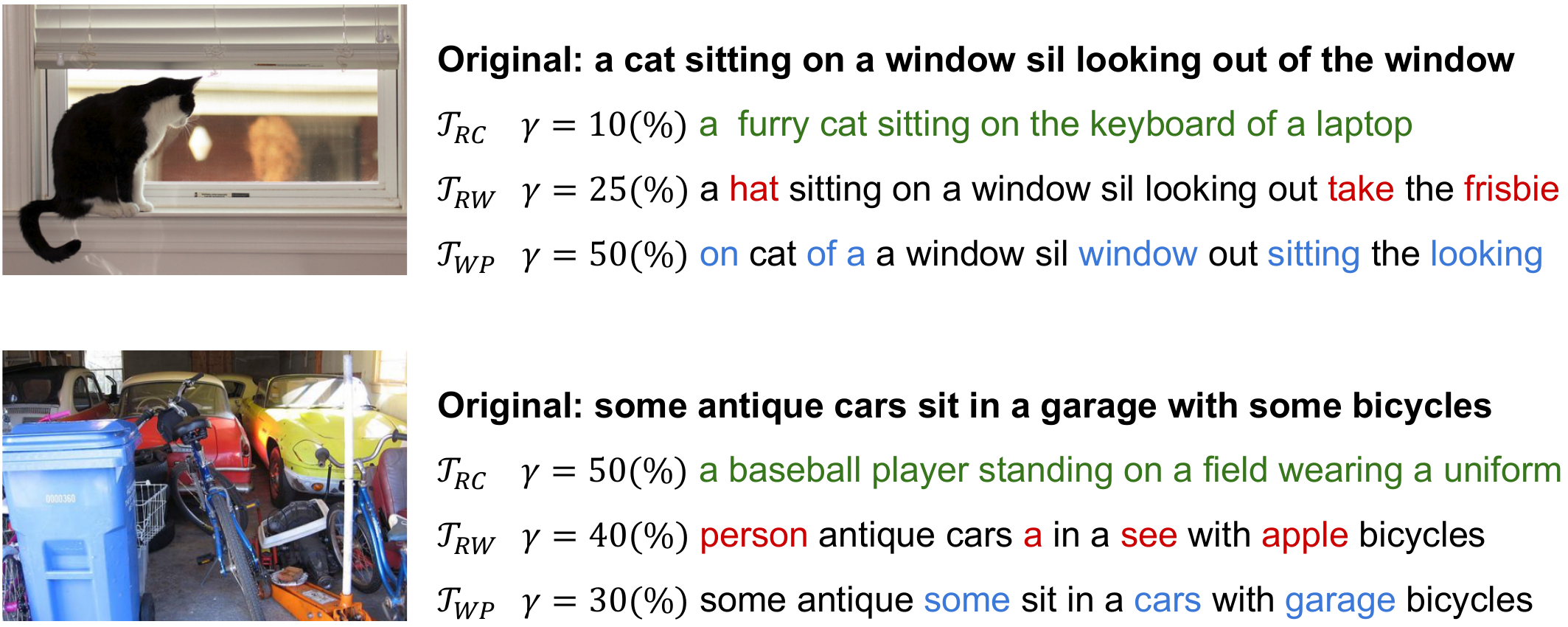}
\end{center}
   \caption{
Example ground truth captions before and after transformations.
Robustness of our learned metrics is evaluated on human captions from similar images ($\mathcal{T}_{RC}$) as well as with random ($\mathcal{T}_{RW}$) and permuted ($\mathcal{T}_{WP}$) words.
}
\label{fig:r_examples}
\end{figure}

\subsection{Data sampling and augmentation}
\label{sec:data_augmentation}
We would like to use data augmentation to incorporate pathological cases as negative examples during training.
We define several transformations of the training data to generate a large amount of pathological sentences.
Formally, a transformation $\mathcal{T}$ takes an image-caption dataset and generates a new one:
\begin{equation}
\mathcal{T}(\{(c,i) \in \mathcal{D}\}; \gamma) = \{(c'_1,i'_1), \dots, (c'_n, i'_n)\}    
\end{equation}
where $i, i'_i$ are images, $c, c'_i$ are captions, $\mathcal{D}$ is a list of caption-image tuples representing the original dataset, and $\gamma$ is a hyper-parameter that controls the strength of the transformation.
Specifically, we define following three transformations to generate pathological image-captions pairs:

\textbf{Random Captions(RC)}.
To ensure our metric pays attention to the image content, we randomly sample human written captions from other images in the training set:
\begin{equation}
\mathcal{T}_{RC}(\mathcal{D}; \gamma) = \{(c', i) | (c,i),(c',i') \in \mathcal{D}, i'\in \mathcal{N}_\gamma(i) \}
\end{equation}
where $\mathcal{N}_\gamma(i)$ represents the set of images that are top $\gamma$ percent nearest neighbors to image $i$.

\textbf{Word Permutation(WP)}. 
To make sure that our metric pays attention to sentence structure, we randomly permute at least 2 words in the reference caption:
\begin{equation}
\mathcal{T}_{WP}(\mathcal{D}; \gamma) = \{(c',i) | (c,i) \in \mathcal{D}, c' \in \mathcal{P}_\gamma(c)\setminus\{c\} \}
\end{equation}
where $\mathcal{P}_\gamma(c)$ represents all sentences generated by permuting $\gamma$ percent of words in caption $c$.

\textbf{Random Word (RW)}. To explore rare words we replace from 2 to all words of the reference caption with random words from the vocabulary:
\begin{equation}
\mathcal{T}_{RW}(\mathcal{D}; \gamma) = \{(c',i) | (c, i) \in \mathcal{D}, c' \in \mathcal{W}_\gamma(c)\setminus\{c\} \}
\end{equation}
where $\mathcal{W}_\gamma(c)$ represents all sentences generated by randomly replacing $\gamma$ percent words from caption $c$.

Note that all the $\gamma$'s are specifically defined to be a percentage. $\gamma\%=0$ denotes the original caption without transformation, while $\gamma\%=1$ provides the strongest possible transformations.
Fig.~\ref{fig:r_examples} shows example captions before and after these transformations.

\begin{figure}[t]
\begin{center}
\includegraphics[width=\columnwidth]{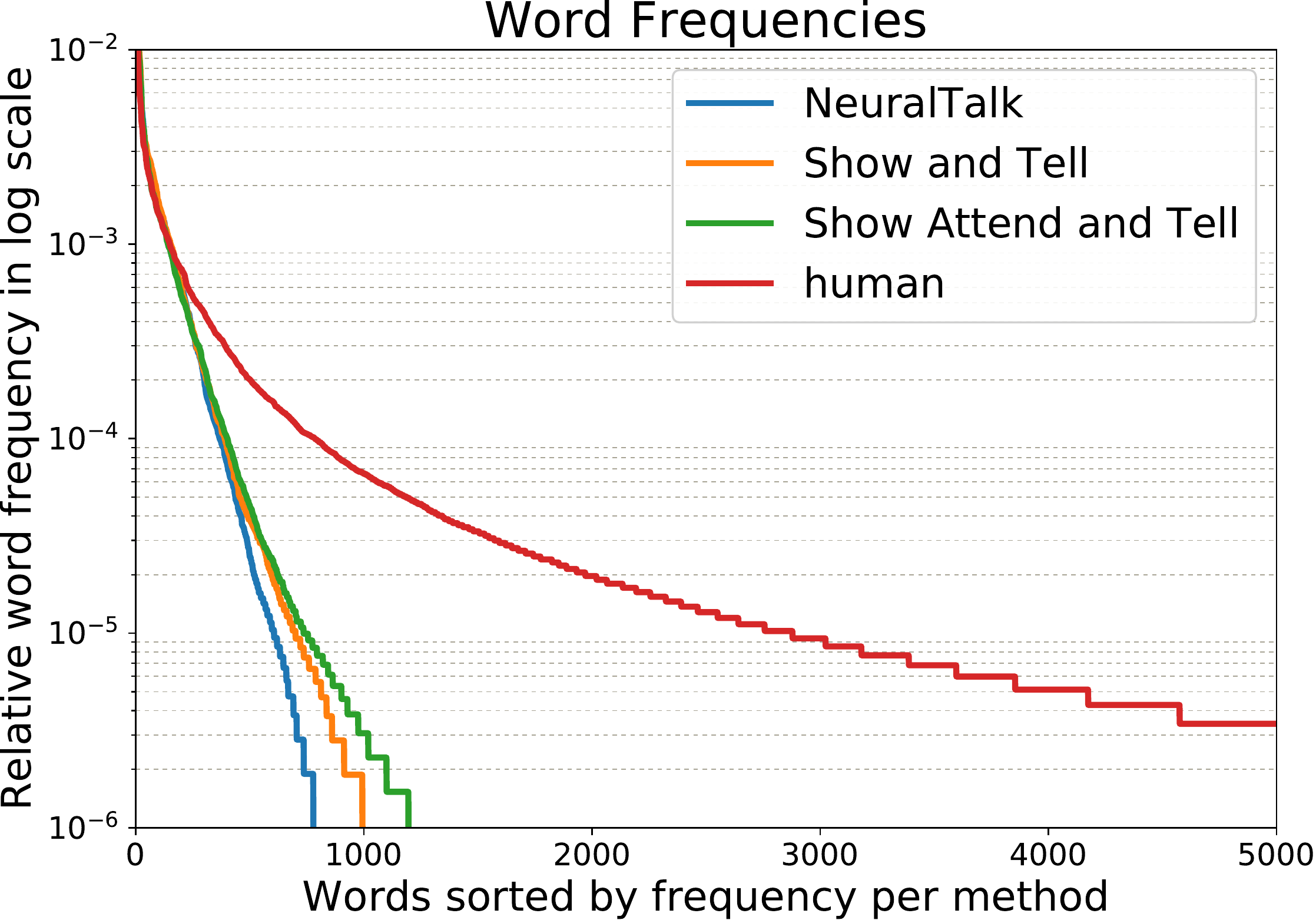}
\end{center}
   \caption{Relative word frequency (in log scale) in the captions generated by ``NeuralTalk'' \cite{karpathy2015deep}, ``Show and Tell'' \cite{showandtell}, ``Show, Attend and Tell'' \cite{showattendtell}, and human captions.
   Machine generated captions have drastically different word frequency distributions from human written captions, as human captions tend to contain much more infrequent words.
   As a result, a discriminator could simply detect the rare words and achieve low classification loss.
}
\label{fig:word_frequency}
\end{figure}

The need for data augmentation can be further illustrated by observing the word frequencies.
Fig.~\ref{fig:word_frequency} shows the relative word frequency in the captions generated by three popular captioning models as well as the frequency in human captions. 
Apparently, a discriminator can easily tell human and generated captions apart by simply looking at what words are used.
In fact, a simple critique only trained on human written and machine-generated captions tends to believe that a sequence of random words is written by a human, simply because it contains many rare words.
To address this problem, our augmented data also includes captions generated using Monte Carlo Sampling, which contains a much higher variety of words.

\subsection{Performance Evaluation}
\label{sec:capability_reliability}
A learned critique should be capable of correctly distinguishing human written captions from machine generated ones.
Therefore, the objective of the critique is to assign scores close to 0 to generated captions and scores close to 1 to human captions.
In light of this, we define the performance of a critique as how close it gets to the ideal objectives, which is either the score assigned to a human caption or one minus the score assigned to a generated caption:
\[
    s(\Theta, (\hat{c},i)) =  
\begin{cases}
    1 - \text{score}_\Theta(\hat{c},i),&  \text{if } \hat{c} \text{ is generated}\\
    \text{score}_\Theta(\hat{c},i),& \text{otherwise}
\end{cases}
\]
where $\Theta$ represents the critique, $\hat{c}$ is the candidate caption, and $i$ is the image $\hat{c}$ summarizes. The performance of a model is then defined as the averaged performance on all the image-caption pairs in a test or validation set:

\begin{equation}
s(\Theta, \mathcal{D}) = \frac{1}{|\mathcal{D}|} \sum_{(\hat{c},i)\in \mathcal{D}} s(\Theta, (\hat{c},i))
\end{equation}
where $\mathcal{D}$ is the set of all image-caption pairs in a held-out validation or test set.

Given a pathological transformation $\mathcal{T}$ and $\gamma$, we could compute the average score of a metric $\Theta$ on the transformed validation set $\mathcal{T}(\mathcal{D},\gamma)$, \ie $s(\Theta, \mathcal{T}(\mathcal{D},\gamma))$. We define the robustness score with respect to transformation $\mathcal{T}$ as the Area-Under-Curve (AUC) of $s(\Theta, \mathcal{T}(\mathcal{D},\gamma))$ by varying all possible $\gamma$:
\begin{equation}
R(\Theta, \mathcal{T}) = \int s(\Theta, \mathcal{T}(\mathcal{D},\gamma)) d\gamma
\end{equation}

We expect a robust evaluation metric to give low scores to the image-caption pairs generated by the pathological transformations.
To compare metrics with different scales, we normalize the scores given by each metric such that the ground truth human caption receives a score of 1.
Detailed experiments are presented in Sec.~\ref{sec:capability} and Sec.~\ref{sec:robustness}.

\subsection{Using the Learned Metrics}
To use the learned metrics in practice, one needs to first fix both the model architecture of the discriminator and all the hyper-parameters of the training process. 
When evaluating a captioning model, we need the generated captions of the model for a set of images (\ie, validation or test set of a image captioning dataset). 
We then split the results into two folds. 
The discriminative metric is trained with image-caption pairs in first fold as training data, together with ground truth captions written by human.
Then we use the trained metric to score the image-caption pairs on the other fold. 
Similarly, we score all the image-caption pairs in the first fold using a metric trained from the second fold. 
Once we get all the image-caption pairs scored in the dataset, the average score will be used as the evaluation of the captioning model.
One could reduce the variance of the evaluation score by training the metric multiple times and use the averaged evaluation score across all the runs.

\section{Experiments}
\label{sec:experiments}


\subsection{Experiment setup}
\label{sec:setup}
\textbf{Data.}
We use the COCO dataset~\cite{coco} to evaluate the performance of our proposed metric.
To test the capability (Sec.~\ref{sec:capability}) and robustness (Sec.~\ref{sec:robustness}) of the proposed models, we use the data split from~\cite{karpathy2015deep}, which re-splits the original COCO dataset into a training set with 113,287 images, a validation and a test set, each contains 5,000 images.
Each image is annotated by roughly 5 human annotators.
We use the validation set for parameter tuning.
For the system level human correlation study (Sec.~\ref{sec:correlation}), we use 12 submission entries from the 2015 COCO Captioning Challenge on the COCO validation set~\footnote{Among 15 participating teams, 3 didn't provide submissions on validation set. Thus, we use submission entries from the remaining 12 teams.}.

The caption level human correlation study (Sec.~\ref{sec:cap-correlation}) uses human annotations in Flickr 8k dataset~\cite{flickr8k}. 
Flickr 8k collects two sets of human annotations, each on a different set of image caption pairs. Among these image-caption pairs, candidate captions are sampled from human captions in the dataset.
In the first set of human annotation (Expert Annotation), human experts are asked to rate the image-caption pairs with scores ranging from 1: The selected caption is unrelated to the image to 4: The selected caption describes the image without any errors.
The second set of annotation (Crowd Flower Annotation) is collected by asking human raters to decide whether a caption describes the corresponding image or not.

\textbf{Image Captioning Models.}
We use publicly available implementations of ``NeuralTalk'' (\textbf{NT}) \cite{karpathy2015deep}, ``Show and Tell''  (\textbf{ST}) \cite{showandtell}, ``Show, Attend and Tell''  (\textbf{SAT}) \cite{showattendtell} as image captioning models to train and evaluate our metric.

\textbf{Implementation Details.}
Our image features are extracted from a Deep Residual Network with 152 layers (ResNet-152)~\cite{resnet} pre-trained on ImageNet.
We follow the preprocessing from \cite{showandtell,showattendtell,karpathy2015deep} to prepare vocabulary on COCO dataset.
We fix the step size of the LSTM to be $15$, padding shorter sentences with a special token while cutting longer ones to 15 words.
All words are represented as 300-dimensional vectors initialized from GloVe \cite{glove}.
We use a batch size of 100 and sample an equal number of positive and negative examples in each batch.
Linear projection is used to reduce the dimension of image feature to match that of caption features.
For Compact Bilinear Pooling, we use the feature dimension of 8192 as suggested in~\cite{cbp}.
We use 1 LSTM layer with a hidden dimension of 512 in all experiments unless otherwise stated.
All the model are trained using Adam~\cite{adam} optimizer for $30$ epochs with an initial learning rate of $10^{-3}$.
We decay the learning rate by a factor of $0.9$ after every epoch.
Our code (in Tensorflow \cite{tensorflow}) is available at: \url{https://github.com/richardaecn/cvpr18-caption-eval}.

\begin{figure}[t]
\centering
\subfigure[scores for human captions]{
\includegraphics[width=\columnwidth]{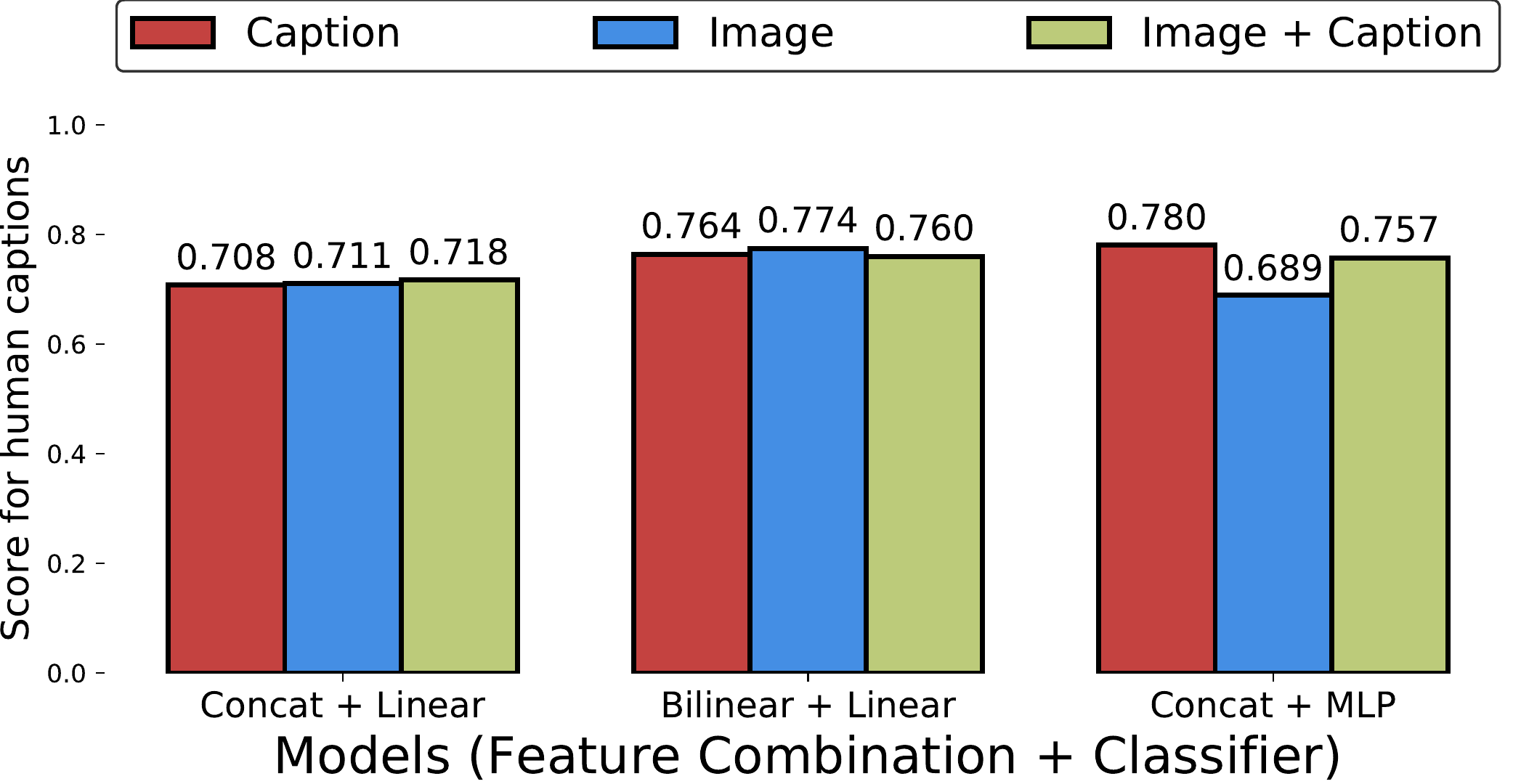}
   \label{fig:capability_human}
 }
\subfigure[scores for generated captions by \textbf{ST}, \textbf{SAT} and \textbf{NT}]{
\includegraphics[width=\columnwidth]{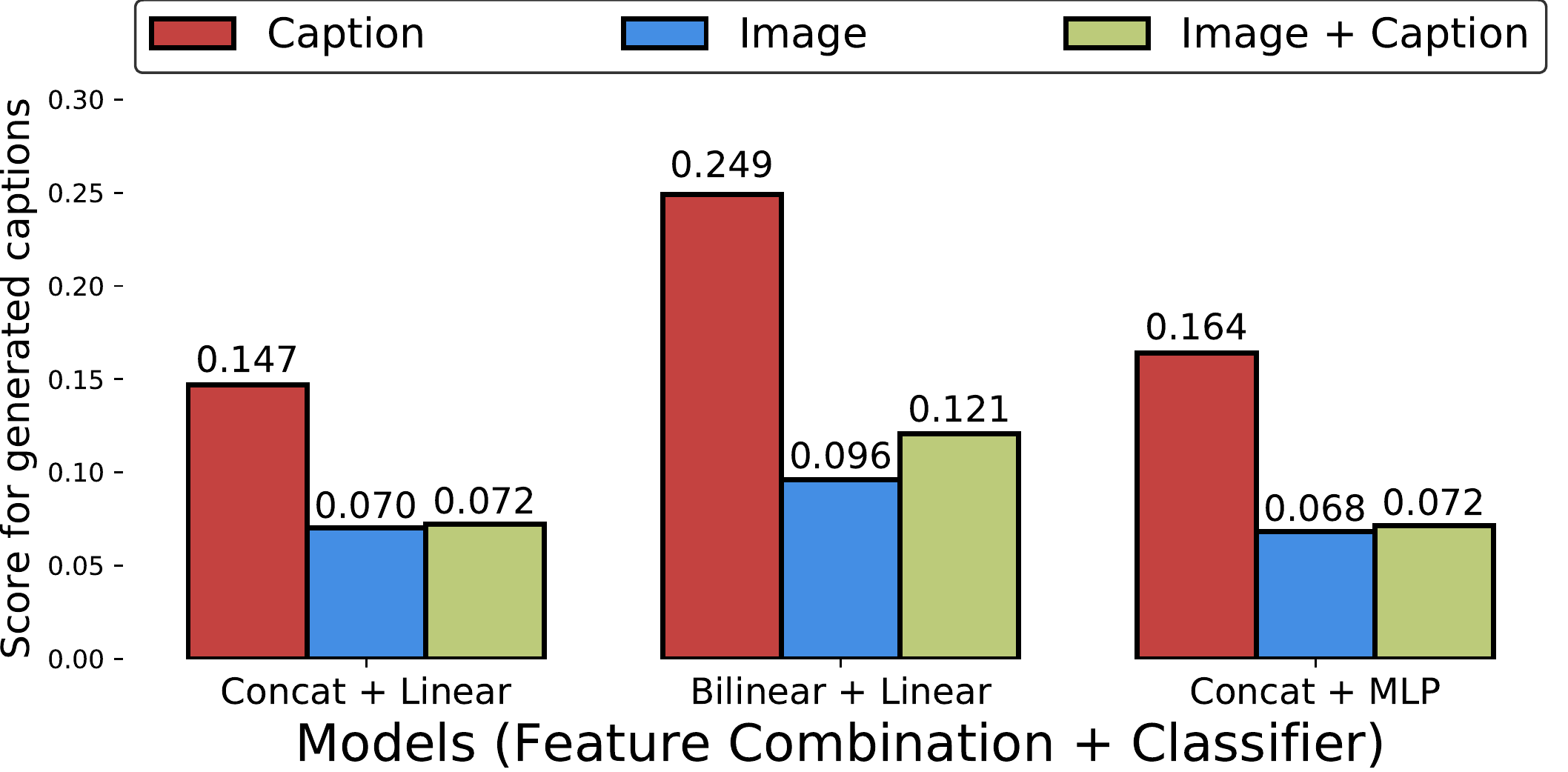}
   \label{fig:capability_gen}
 }
\caption{
Top: Average score of human captions from the validation set. Bottom: Average score of generated captions.
Color of the bar indicates what context information is used for the critique.
The horizontal axis represents three different strategies for the combination of the features from candidate caption with the context as well as the used classifier (Concat + Linear: concatenation and linear classifier; Bilinear + Linear: compact bilinear pooling and linear classifier; Concat + MLP: concatenation and MLP with one hidden layer of size 512).
} 
\label{fig:capability}
\end{figure} 

\begin{figure*}[h!]
\centering
\subfigure{
\includegraphics[width=0.32\linewidth]{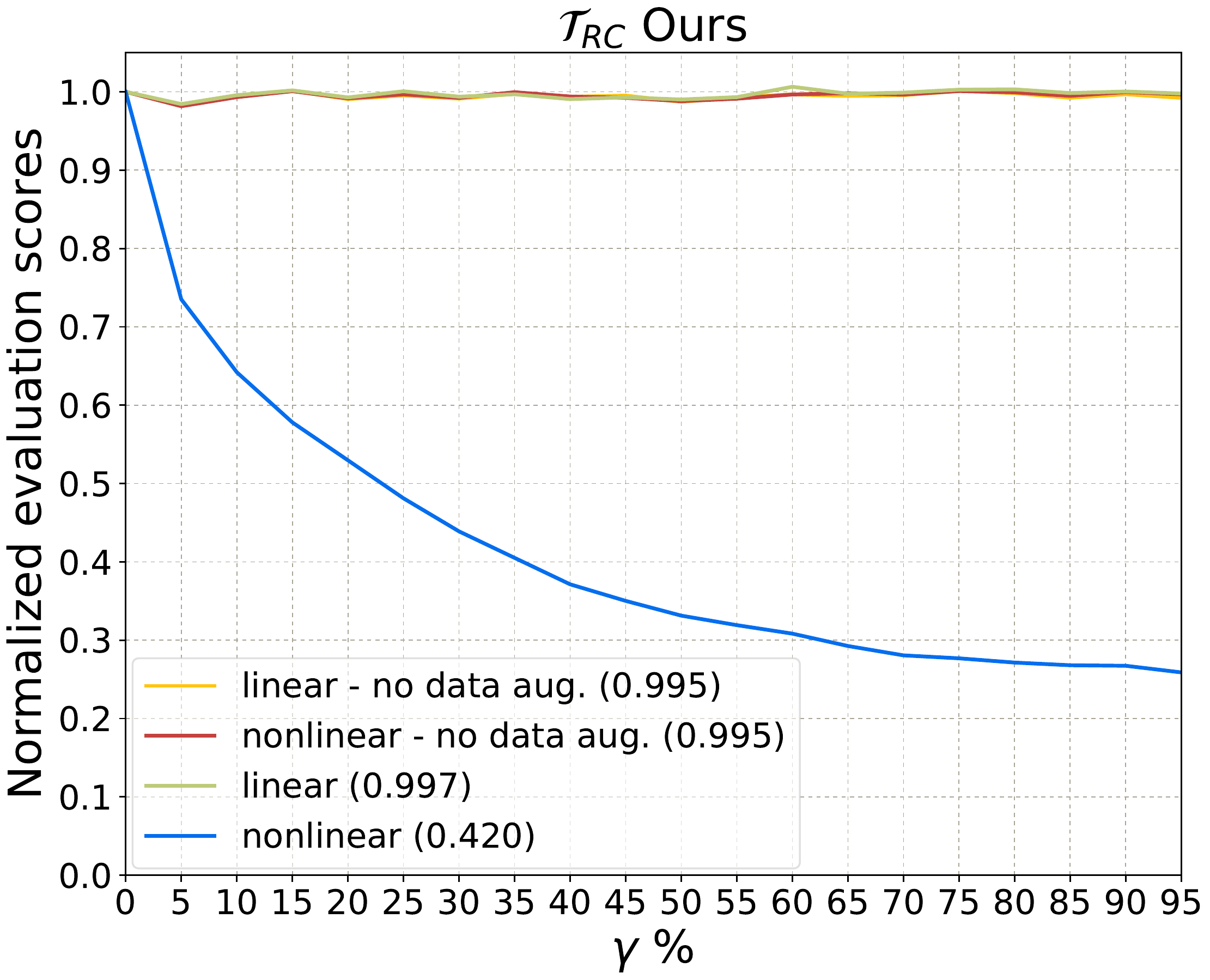}
  \label{fig:a}
 }
\subfigure{
\includegraphics[width=0.32\linewidth]{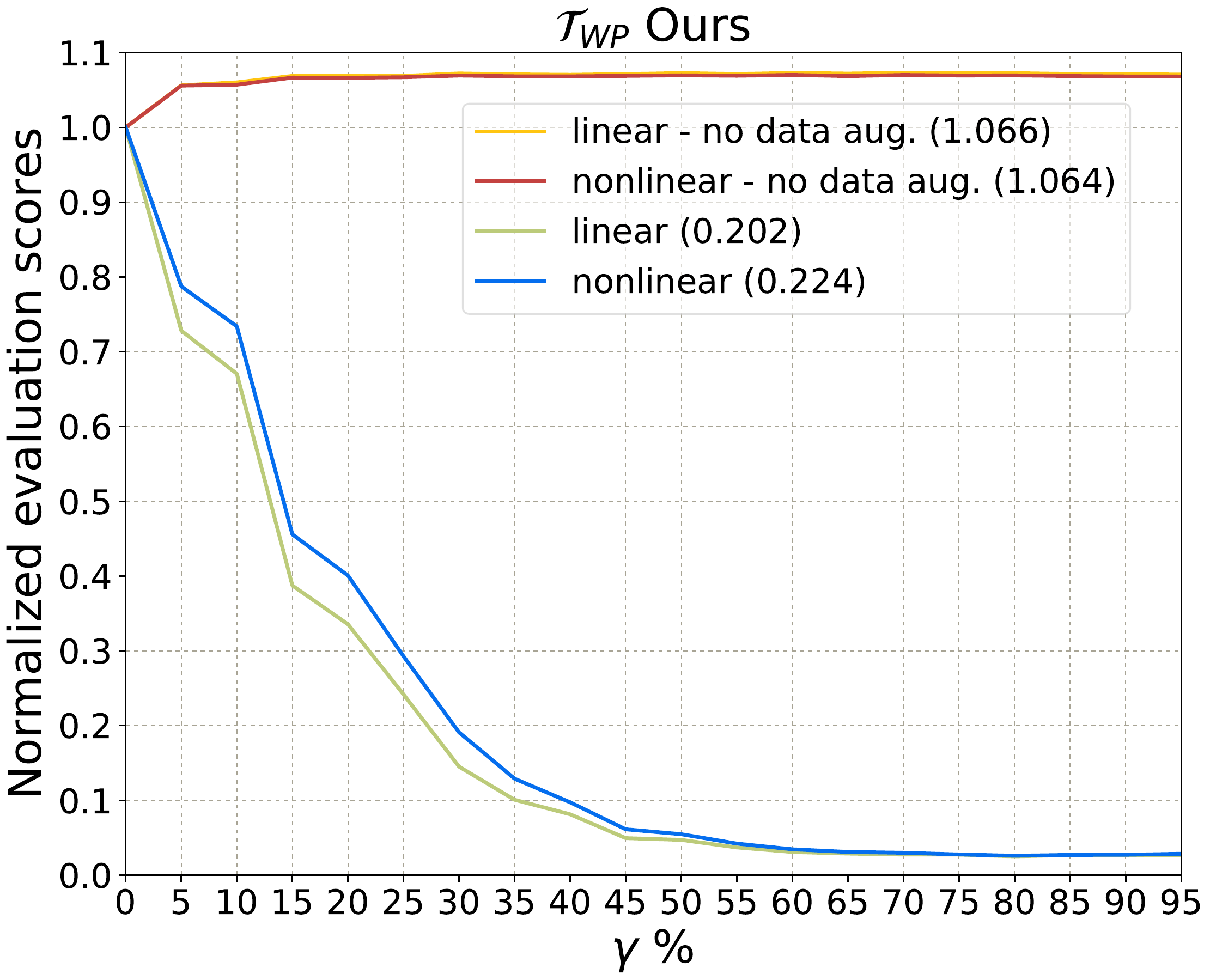}
  \label{fig:b}
 }
\subfigure{
\includegraphics[width=0.32\linewidth]{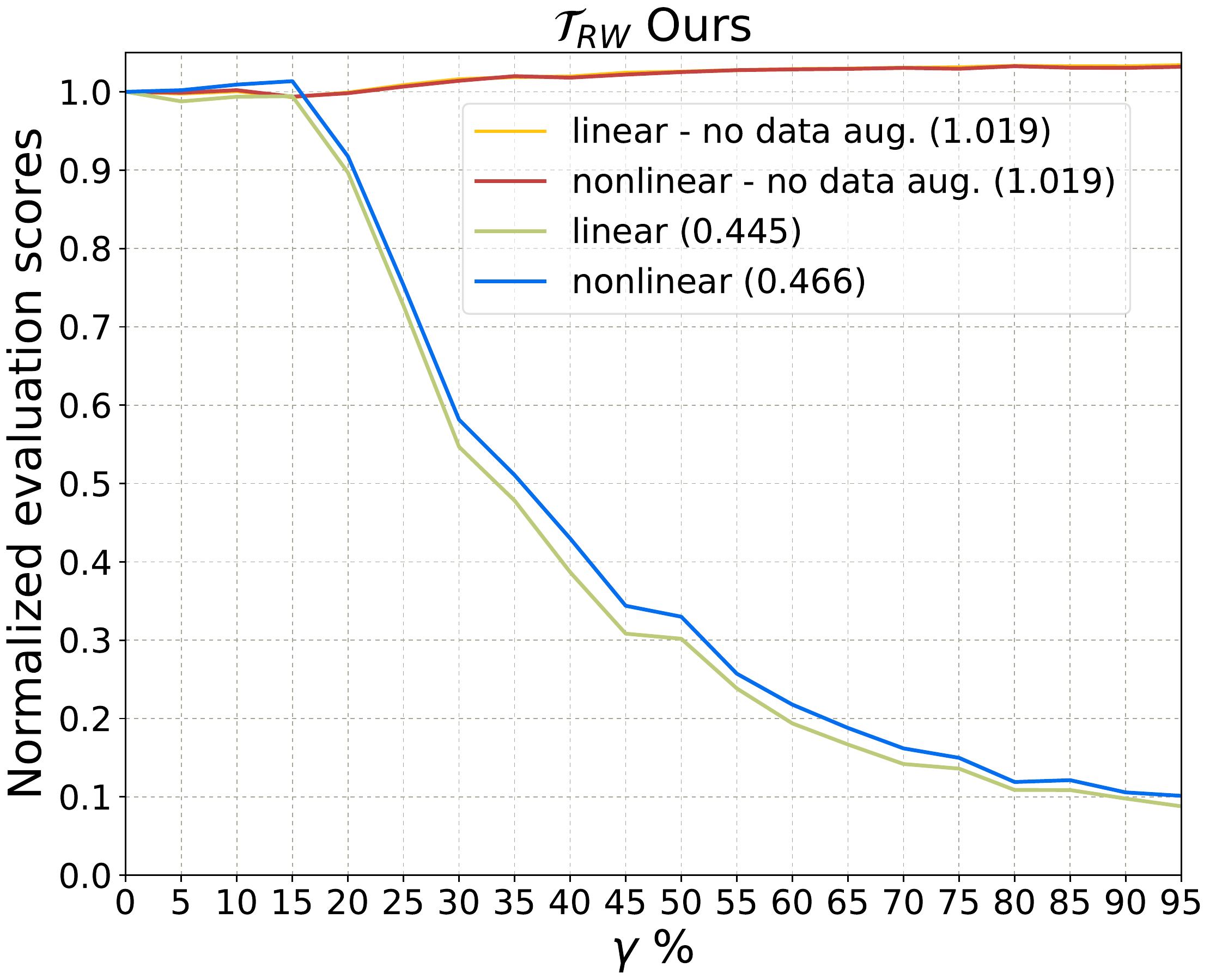}
  \label{fig:c}
 }
\subfigure{
\includegraphics[width=0.32\linewidth]{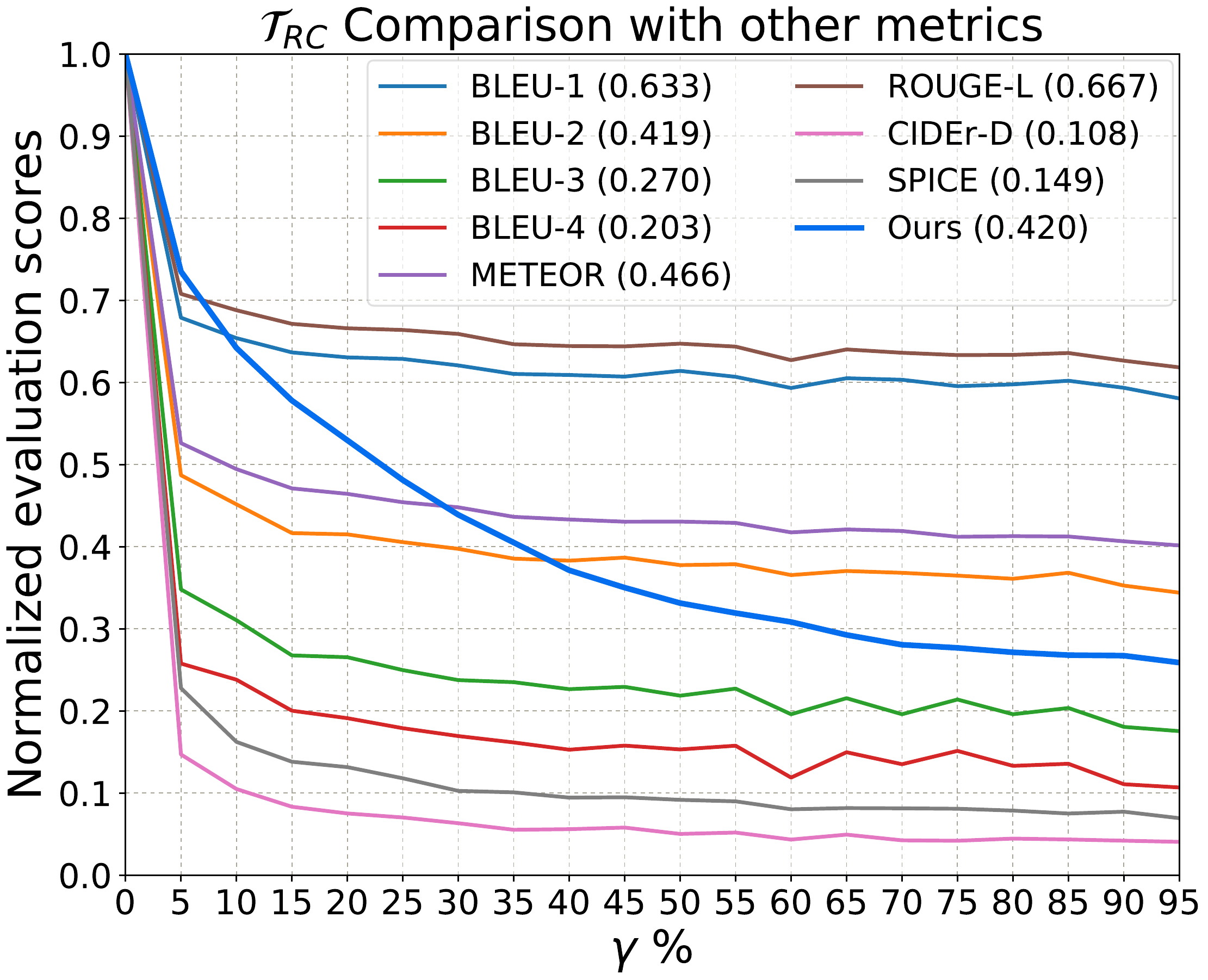}
  \label{fig:d}
 }
\subfigure{
\includegraphics[width=0.32\linewidth]{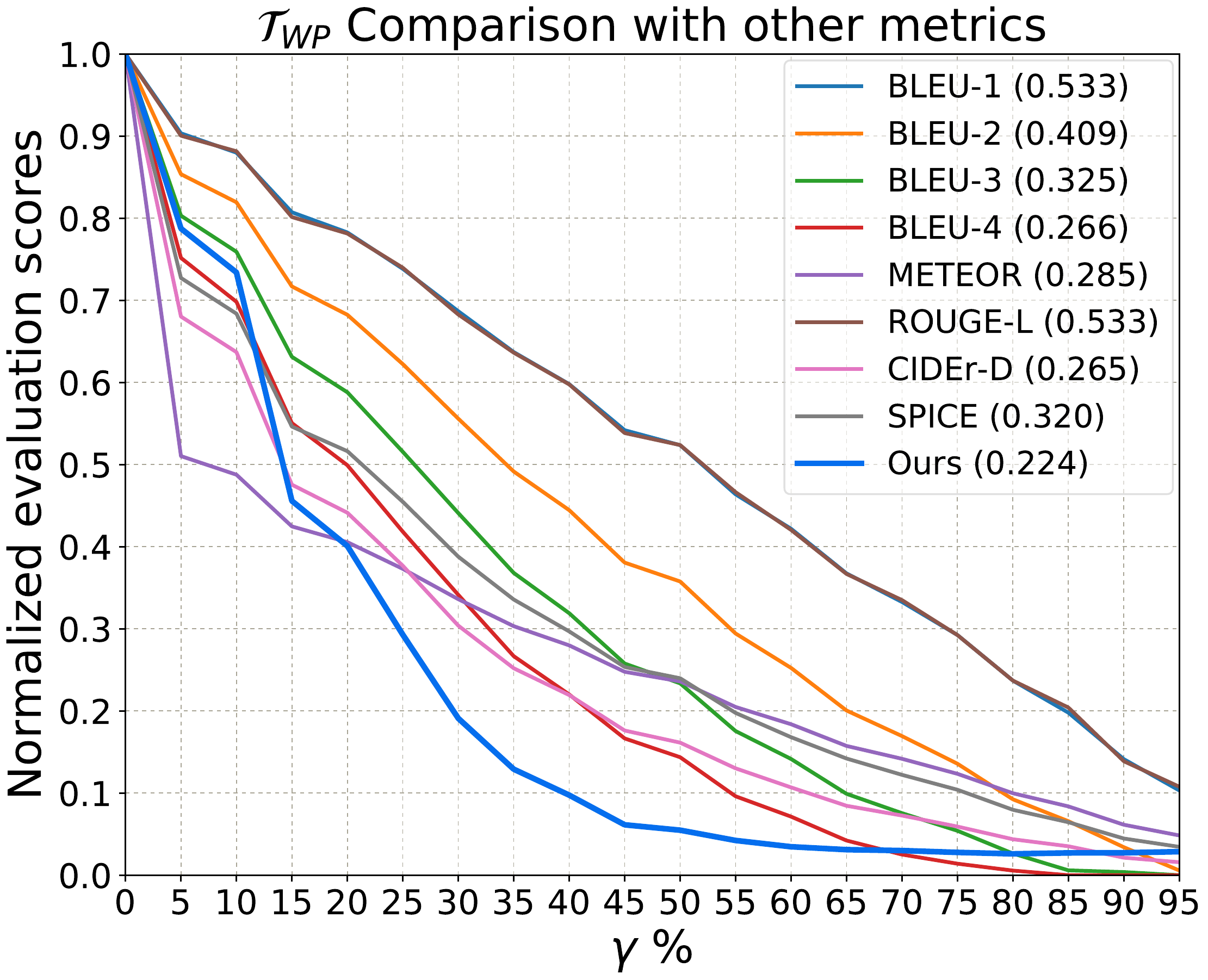}
  \label{fig:e}
 }
\subfigure{
\includegraphics[width=0.32\linewidth]{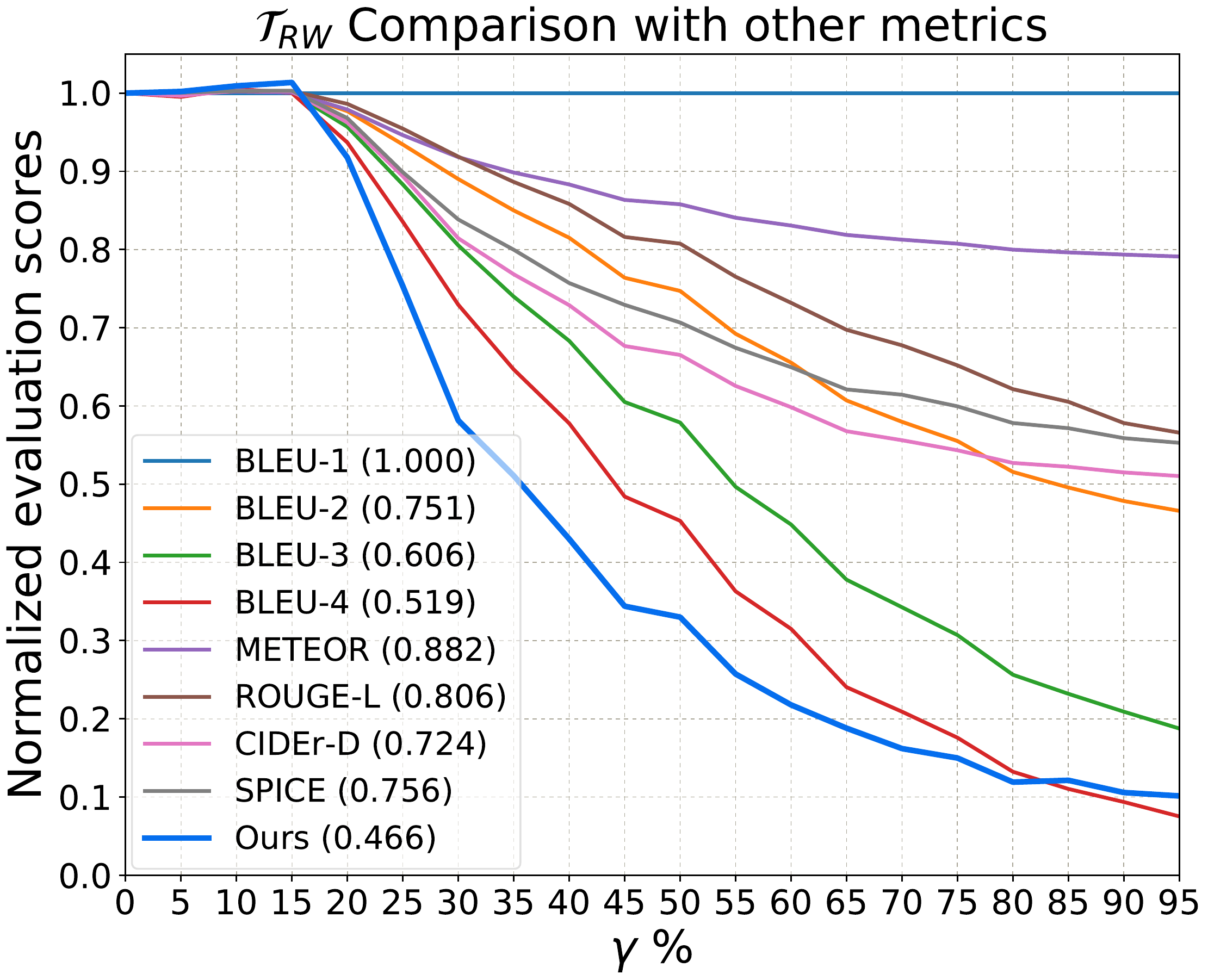}
  \label{fig:f}
 }
\caption{
Normalized evaluation score for transformations $\mathcal{T}_{RC}$ (human caption from other images), $\mathcal{T}_{WP}$ (random permutation of words), and $\mathcal{T}_{RW}$ (words replaced by random words) with different amount of transformation ($\gamma\%$).
When $\gamma\%=0\%$, the original dataset is kept unchanged; when $\gamma\%=100\%$, maximum amount of transformation is applied to the dataset.
The first row shows results of our metrics using either linear or non-linear model trained with or without data augmentation. The second row compares our non-linear model trained with data augmentation to other metrics.
The score after each metric shows the robustness score defined in Sec.~\ref{sec:capability_reliability}, \ie,\ the Area Under Curve (AUC).
The lower the score the more robust the metric is.
}
\label{fig:robustness}
\end{figure*}

\subsection{Capability}
\label{sec:capability}
To measure the capability of our metric to differentiate between human and generated captions, we train variants of models using generated captions from \textbf{ST}, \textbf{SAT} and \textbf{NT}, together with human captions from the training set.
Fig.~\ref{fig:capability_human} and Fig.~\ref{fig:capability_gen} show the average score on the validation set for human and generated captions respectively.
The results show that all models give much higher scores to human captions than machine generated captions, indicating that they are able to differentiate human-written captions from the machine-generated ones.

\vspace{-0.5mm}
With respect to the choice of context, we observe that including image features into the context clearly improves performance. Also, adding a reference caption does not lead to a significant improvement over only using image features. This indicates that the image itself provides enough contextual information for the critique to successfully discriminate between human and machine generated captions.
The reason that none of the commonly used metrics includes images as context is likely due to the difficulties of capturing image-text similarity.
Our metric circumvents this issue by implicitly learning the image-text relationship directly from the data.

It is worth noting that achieving high model performance in terms of discrimination between human and generated captions does not necessarily imply that the learned metric is good.
In fact, we observe that a critique trained without data augmentation can achieve even higher discrimination performance.
Such critique, however, also gives high scores to human written captions from other images, indicating that the classification problem is essentially reduced to putting captions into categories of human and non-human written without considering the context image.
If trained with the proposed data sampling and augmentation technique, the critique learns to pay attention to image context.

\subsection{Robustness}
\label{sec:robustness}
To evaluate whether the proposed metric can capture pathological image-caption pairs, we conduct robustness studies as described in Sec.~\ref{sec:capability_reliability} on the three pathological transformations defined in Sec.~\ref{sec:data_augmentation}.
The robustness comparisons are illustrated in Fig.~\ref{fig:robustness}.
In the first row we compare different variants of the proposed metric.
The results illustrate that, although achieving high discrimination performance, a metric learned without data sampling or augmentation also gives high scores to human captions from other images ($\mathcal{T}_{RC}$), with random words ($\mathcal{T}_{RW}$), or word permutations ($\mathcal{T}_{WP}$).
This indicates that the model tends to focus on an overall human vs. non-human classification without considering contextual information in the image or the syntactic structure of the candidate sentence.

Further, even with data augmentation, a linear model with concatenated context and candidate caption features gives high scores to human captions from other images, possibly because there is no sufficient interaction between the context and candidate caption features. 
Non-linear interactions such as Compact Bilinear Pooling or a non-linear classifier with hidden layers solve this limitation.
The nonlinear model in Fig.~\ref{fig:robustness} refers to a model with concatenated context and candidate features followed by a nonlinear classifier. Compact bilinear pooling (not shown in the figure for clarity of visualization) achieves similar results.

In the second row of Fig.~\ref{fig:robustness} we compare our metric with other commonly used image captioning metrics. 
The proposed metric outperforms all others with respect to random ($\mathcal{T}_{RW}$) as well as permuted ($\mathcal{T}_{WP}$) words and is reasonably robust to human captions from similar images ($\mathcal{T}_{RC}$).
Further, we observe that the recently proposed metrics CIDEr and SPICE perform well for human captions from similar images, but fall behind with respect to sentence structure.
This could be caused by their increased focus on informative and scene specific words.

\subsection{Caption Level Human Correlation}
\label{sec:cap-correlation}
We use both the Expert Annotations and the Crowd Flower Annotations from Flickr 8k dataset~\cite{flickr8k} to compute caption level correlation with human judgments.
We follow the procedure in SPICE paper~\cite{anderson2016spice} to compute the Kendall's $\tau$ rank correlation in the Expert Annotations.
The $\tau$ correlation for the Crowd Flower Annotation is computed between scores generated by the evaluation metric and percentage of raters who think that the caption describes the image with possibly minor mistakes.
During training, all negative samples are generated by transformation $\mathcal{T}_{RC}$,~\ie, human caption from random image. 

The results in Table~\ref{tab:flickr-correlation} show that our metrics achieve the best caption level correlation in both Expert Annotations and Crowd Flower Annotations.
Note that the Crowd Flower Annotations use a binary rating setup, while the set-up from the Expert Annotations makes a finer-grained ratings.
Despite the fact that our model is trained on a simpler binary objective, it still correlates well with human judgments from the Expert Annotations.
Note that we do not use any human annotations during the training, since all of our negative examples could be generated automatically.

\subsection{System Level Human Correlation}
\label{sec:correlation}
We compare our metric with others on the Pearson's $\rho$ correlation between all common metrics and human judgments collected in the 2015 COCO Captioning Challenge~\cite{cocochallenge}. 
In particular, we use two human judgment M1: Percentage of captions that are evaluated as better or equal to human caption and M2: Percentage of captions that pass the Turing Test.
We don't use M3: correctness, M4: detailness and M5: salience, as they are not used to rank image captioning models, but are intended for an ablation study to understand which aspects make captions good.

Since we don't have access to the COCO test set annotations, where the human judgments are collected on, we perform our experiments on the COCO validation set. 
There are 15 teams participated in the 2015 COCO captioning challenge and we use 12 of them that submitted results on the validation set.
We assume the human judgment on the validation set is sufficiently similar to the judgment on the test set. 
We don't use any additional training data besides the submission files on the validation set and the data augmentation described in Sec.~\ref{sec:data_augmentation}.
To get evaluation scores on the whole validation set, we split the set in two halves and, for each submission, train our critique on each split and get scores (probability of being human written) on the other.

\begin{table}[t]
\begin{center}
\begin{tabular}{@{}ccc@{}}
\hline
\multicolumn{1}{c|}{}          & \multicolumn{1}{c|}{\textbf{Expert Annotations}}   & \multicolumn{1}{c}{\textbf{Crowd Flower}}   \\
\hline
\multicolumn{1}{c|}{BLEU-1}    & \multicolumn{1}{c|}{0.191*}    & \multicolumn{1}{c}{0.206}          \\
\multicolumn{1}{c|}{BLEU-2}    & \multicolumn{1}{c|}{0.212}    & \multicolumn{1}{c}{0.212}          \\
\multicolumn{1}{c|}{BLEU-3}    & \multicolumn{1}{c|}{0.209}    & \multicolumn{1}{c}{0.204}          \\
\multicolumn{1}{c|}{BLEU-4}    & \multicolumn{1}{c|}{0.206*}    & \multicolumn{1}{c}{0.202}          \\
\multicolumn{1}{c|}{METEOR}    & \multicolumn{1}{c|}{0.308*}    & \multicolumn{1}{c}{0.242}          \\
\multicolumn{1}{c|}{ROUGE-L}   & \multicolumn{1}{c|}{0.218*}    & \multicolumn{1}{c}{0.217}          \\
\multicolumn{1}{c|}{CIDEr}     & \multicolumn{1}{c|}{0.289*}    & \multicolumn{1}{c}{0.264}          \\
\multicolumn{1}{c|}{SPICE}     & \multicolumn{1}{c|}{0.456}    & \multicolumn{1}{c}{0.252}          \\
\multicolumn{1}{c|}{\textbf{Ours}}      & \multicolumn{1}{c|}{\textbf{0.466}}    & \multicolumn{1}{c}{\textbf{0.295}}          \\
\multicolumn{1}{c|}{Inter-human}&\multicolumn{1}{c|}{0.736}    & -              \\ 
\hline
\multicolumn{3}{l}{\textbf{Expert Annotations}: experts score image-caption pairs} \\
\multicolumn{3}{l}{from 1 to 4; 1 means caption doesn't describe the image.}           \\ \midrule
\multicolumn{3}{l}{\textbf{Crowd Flower}: human raters mark 1 if the candidate}  \\
\multicolumn{3}{l}{caption describes the image, and mark 0 if not.}                  \\
\hline
\end{tabular}
\end{center}
\caption{Caption level Kendall's $\tau$ correlation between Flickr 8K \cite{flickr8k}'s human annotations and evaluation metrics' scores. Our reported scores with * differ from the ones reported in SPICE~\cite{anderson2016spice}.
}
\label{tab:flickr-correlation}
\end{table}

\begin{table}[t]
\begin{center}
\begin{tabular}{ c|cc|cc } 
\hline
 & \multicolumn{2}{c|}{\textbf{M1}} & \multicolumn{2}{c}{\textbf{M2}} \\ \cline{2-5}
 & $\rho$ & $p$-value & $\rho$ & $p$-value \\
\hline
BLEU-1 & 0.124 & (0.687) & 0.135 & (0.660) \\
BLEU-2 & 0.037 & (0.903) & 0.048 & (0.877) \\
BLEU-3 & 0.004 & (0.990) & 0.016 & (0.959) \\
BLEU-4 & -0.019 & (0.951) & -0.005 & (0.987) \\
METEOR & 0.606 & (0.028) & 0.594 & (0.032) \\
ROUGE-L & 0.090 & (0.769) & 0.096 & (0.754) \\
CIDEr & 0.438 & (0.134) & 0.440 & (0.133) \\
SPICE & 0.759 & (0.003) & 0.750 & (0.003) \\
\textbf{Ours (no DA)} & \textbf{0.821} & (0.000) & \textbf{0.807} & (0.000) \\
\textbf{Ours} & \textbf{0.939} & (0.000) & \textbf{0.949} & (0.000) \\
\hline
\multicolumn{5}{l}{\textbf{M1}: Percentage of captions that are evaluated as better} \\
\multicolumn{5}{l}{or equal to human caption.}\\
\hline
\multicolumn{5}{l}{\textbf{M2}: Percentage of captions that pass the Turing Test.}\\
\hline
\end{tabular}
\end{center}
\caption{
Pearson's $\rho$ correlation between human judgments and evaluation metrics.
The human correlation of our proposed metric surpasses all other metrics by large margins.
Scores reported in SPICE~\cite{anderson2016spice} were calculated on the COCO test set for all 15 teams, whereas ours were from 12 teams on the COCO validation set.
}
\label{tab:correlation}
\end{table}

\begin{figure}[t]
\begin{center}
\includegraphics[width=\columnwidth]{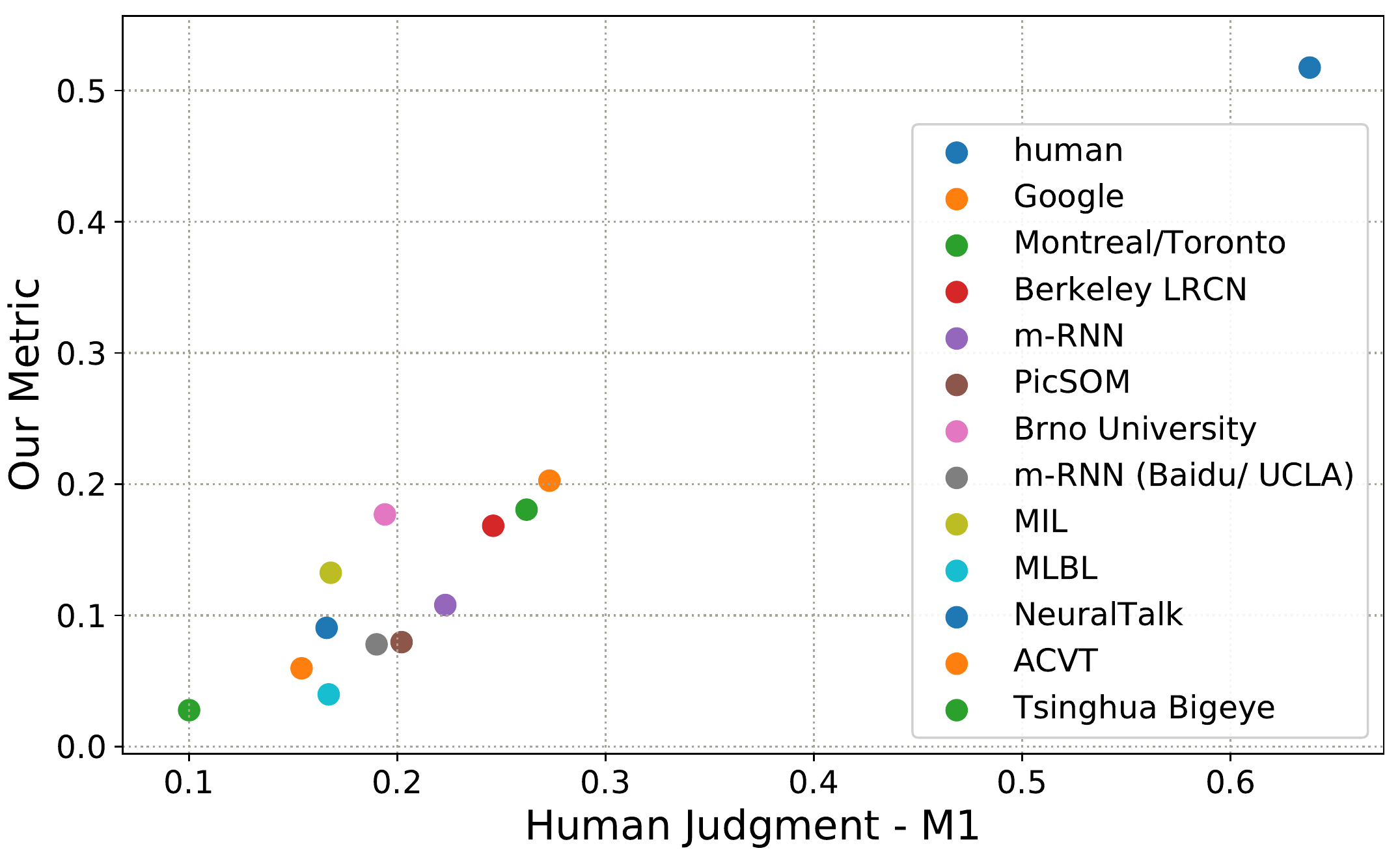}
\end{center}
\caption{Our metric \vs human judgment on COCO validation set.
Our metric is able to reflect most of the rankings from human judgment correctly, especially for top performing methods.}
\label{fig:scatter_plot}
\end{figure}

The results in Table~\ref{tab:correlation} show that our learned metric surpasses all other metrics including the recently proposed SPICE~\cite{anderson2016spice} by large margins, especially trained with data augmentation.
This indicates that aligning the objective with human judgments and using data augmentation yield a better evaluation metric.
Fig.~\ref{fig:scatter_plot} illustrates our metric compared with human judgment - M1 on COCO validation set.
Our metric aligns well with human judgment, especially for top performing methods.

\section{Conclusion and Future Work}
In this paper, we have proposed a novel learning based evaluation metric for image captioning that is trained to act like a human critique to distinguish between human-written and machine-generated captions while also being flexible to adapt to targeted pathological cases.
Further, we have shown how to use data sampling and augmentation to successfully train a metric that behaves robustly against captions generated from pathological transformations.
From extensive experimental evaluations, we have demonstrated that the proposed metric is robust and correlates better to human judgments than previous metrics.
In conclusion, the proposed metric could be an effective complementary to the existing rule-based metrics, especially when the pathological cases are easy to generate but difficult to capture with traditional hand-crafted metrics.

In this study, we have not taken different personalities among human annotators into consideration.
Different human personalities could give rise to different types of human captions. 
One direction of future work could aim to capture the heterogeneous nature of human annotated captions and incorporate such information into captioning evaluation.
Another direction for future work could be training a caption generator together with the proposed evaluation metric (discriminator) in a generative adversarial setting.
Finally, gameability is definitely a concern, not only for our learning based metric, but also for other rule-based metrics.
Learning to be more robust to adversarial examples is also a future direction of learning based evaluation metrics.

\vspace{5pt}
\par\noindent\textbf{Acknowledgments.}
This work was supported in part by a Google Focused Research Award, AWS Cloud Credits for Research and a Facebook equipment donation.
We would like to thank the COCO Consortium for agreeing to run our code on entries in the 2015 COCO Captioning Challenge.

{\small
\bibliographystyle{ieee}
\bibliography{references.bib}
}

\newpage 
\begin{appendices}
\section{Implementation Details}

\subsection{Image Representations}
To extract image features, we use a 152-layer Residual Network (ResNet-152) \cite{resnet} pretrained on ImageNet, which achieved state-of-the-art performance on the large-scale image classification task \cite{imagenet_challenge}. Instead of the standard feature extraction procedure of extracting features from a resized and cropped $224\times 224$ image, we extract the features from the original image without any resizing and cropping. The feature map from the last convolution layer is average-pooled, resulting in a $2048$-dimensional feature vector as our image feature representation. The image features are remain fixed during training.

\subsection{Caption Representations}
We construct a vocabulary list by taking the 10,000 most frequent words that appear at least 5 times in the human annotated captions from the training set. A special token is added to the vocabulary to represent any word that is not among the top 10,000 words. Suppose the length of the vocabulary list is $n$. Each word in the vocabulary can be represented by a one-hot vector $\mathbf{w} \in \{0,1\}^n$, where for word $i$, $w_i=1$ and for all $j \neq i$, $w_j = 0$. Then, a word embedding matrix $E \in \mathbb{R}^{n\times d}$ is used to encode each word as a $d$-dimensional vector $\mathbf{x}=\mathbf{w}E \in \mathbb{R}^d$ as the input to the LSTM. The word embedding is initialized from GloVe \cite{glove}. We use a word embedding dimension of $d=300$ for all of our experiments. We fix the step size of the LSTM to be 15. That is, shorter sentences are padded with a special token and longer captions are cut at 15 words. During training, a mask is applied to remove the padded part of a caption when we compute the classification loss.

\subsection{Training}
During training, we sample equal number of positive and negative examples. To generate positive examples, we first randomly choose an image from the database, and such image should correspond to several reference captions. We use one reference caption as the context, and a different one as the candidate caption. To compose a negative example, we first choose with equal probability one of the following types of negative examples: 1) using a caption generator; 2) sample a caption from a pathologically transformed dataset; or 3) generate a caption using Monte Carlo Sampling. If we are using a pathologically transformed dataset, we will choose in equal probability among three transformations: $\mathcal{T}_{RC}$ (human caption for a different image), $\mathcal{T}_{WP}$ (reference caption with word permutation), and $\mathcal{T}_{RW}$ (reference caption with random word replacement).

\subsection{Evaluation}
To evaluate how good a candidate caption is, we iterate through all the reference captions for the image and compute a score using each reference caption as context for the candidate caption. The average of these scores is the final score for the candidate caption.

To evaluate a caption generator, we train our model for 10 epochs using only this generator to produce the first type of negative examples. We use pathological transformation and Monte Carlo Sampling for all model evaluation. Finally, we use our model to score all candidate captions this generator produces on a held-out set of data. The average of these score is used as the final indicator for how good the caption generator is.

While computing the caption level correlation with human, we first use a candidate metric to compute a score for each pair of image and candidate caption $(i, c)$, where $i$ indicates the image and $c$ indicates the candidate caption. Suppose a $(i,c)$ pair has corresponding human annotations $\mathcal{A}_{i,c}$ and our computed scores $\mathcal{S}_{i,c}$, we create all pairs between human annotations and computed scores $[(h, s) | h \in \mathcal{A}_{i,c}, s \in \mathcal{S}_{i,c}]$. Finally, we compute the Kendall’s $\tau$ Rank Correlation for all score pairs we could generate, \ie., 
\begin{equation}
\tau([(h, s) | h \in \mathcal{A}_{i,c}, s \in \mathcal{S}_{i,c},\forall (i,c)])
\end{equation}


\begin{figure}[h!]
\centering
\subfigure[models with different LSTM layers]{
\includegraphics[width=\linewidth]{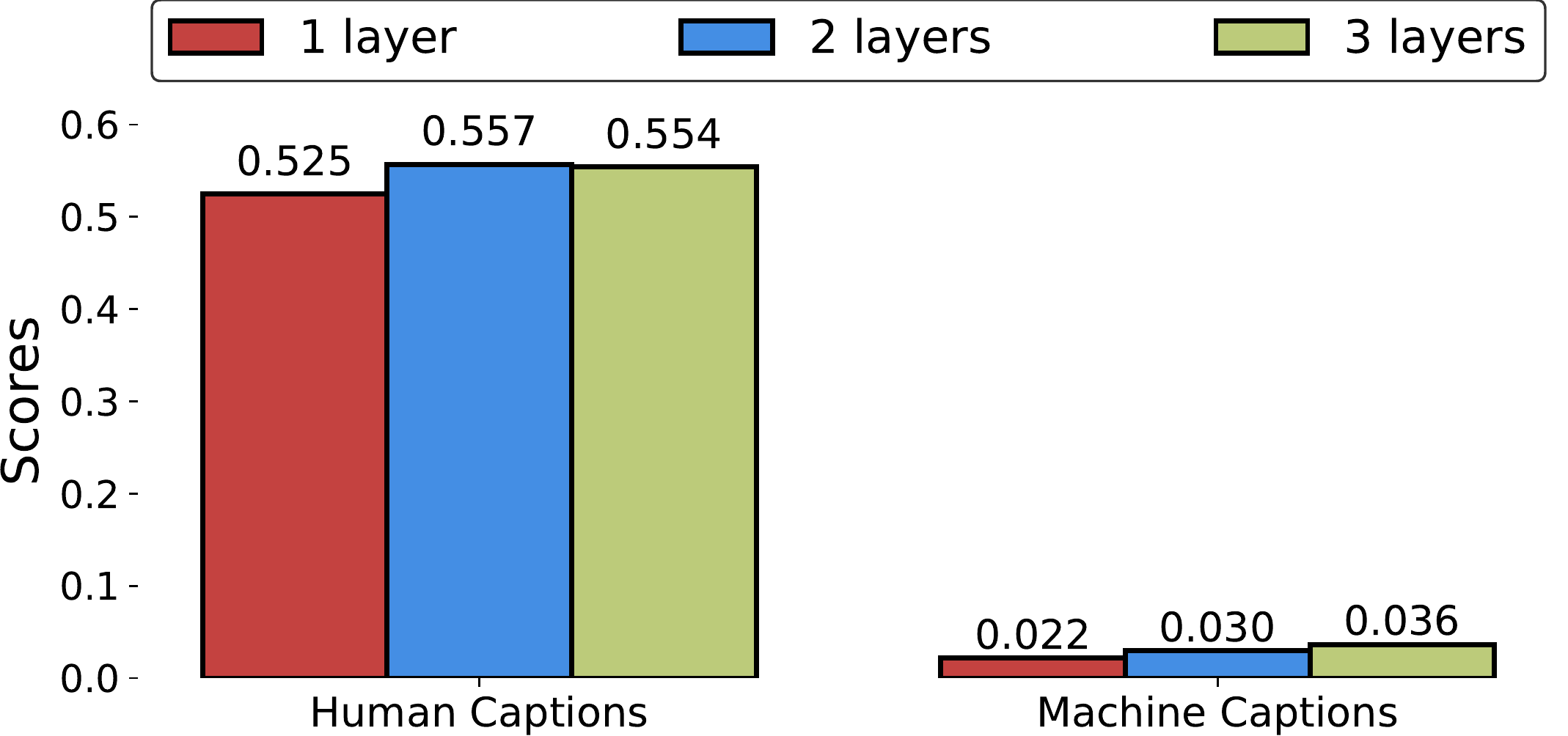}
   \label{fig:1_1}
 }
\subfigure[models with different LSTM hidden feature size]{
\includegraphics[width=\linewidth]{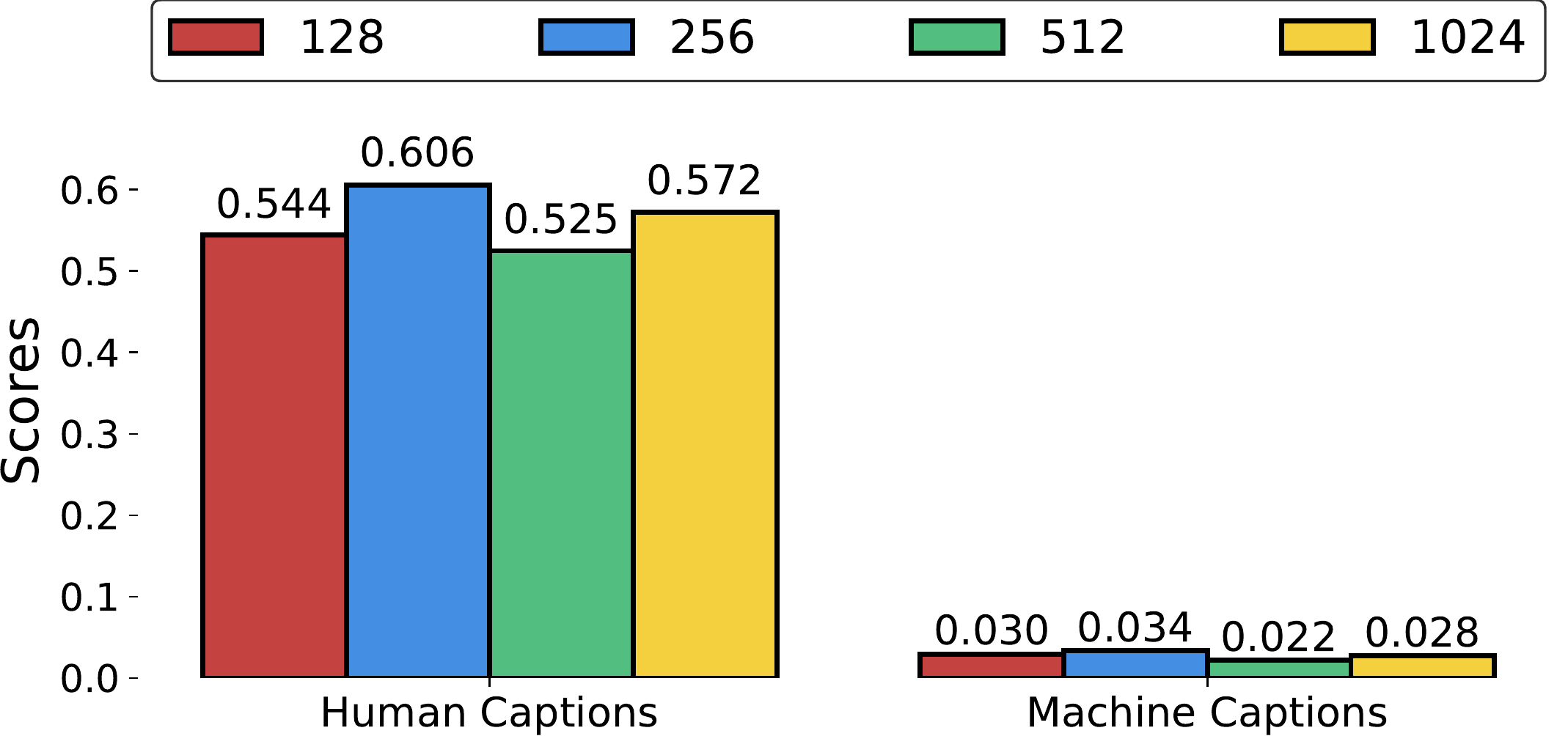}
   \label{fig:1_2}
 }
\caption{
Top: models with variant LSTM layers (512 hidden size). Bottom: models with variant LSTM hidden feature size (1 layer). 
All the models are trained with both image and reference ground truth captions as contexts, using concatenation of context information and candidate caption followed by a linear classifier and with data augmentation.
} 
\label{fig:1}
\end{figure} 

\begin{figure}[h!]
\centering
\subfigure[scores for human captions]{
\includegraphics[width=\linewidth]{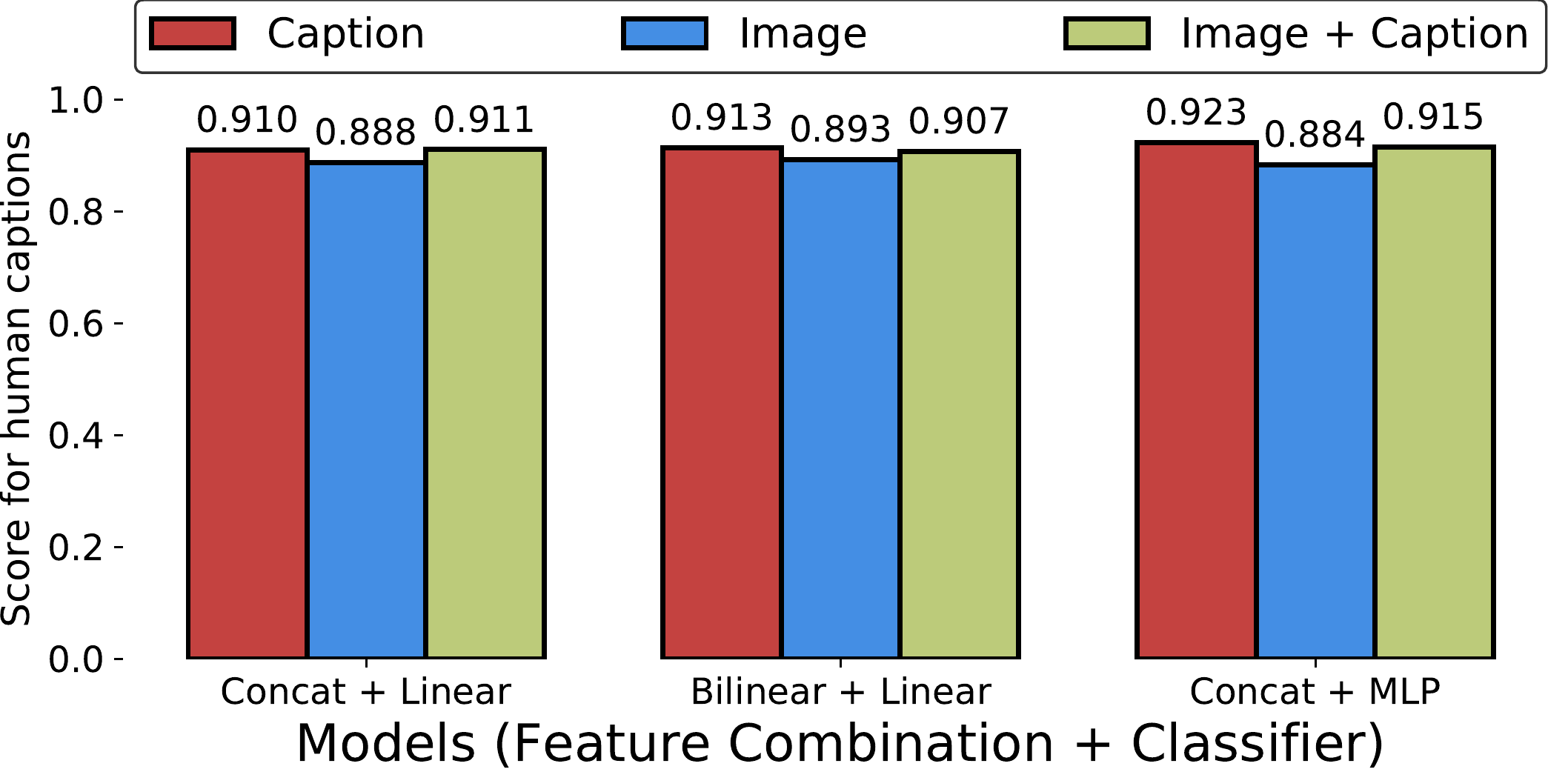}
   \label{fig:2_1}
 }
\subfigure[scores for generated captions by \textbf{ST}, \textbf{SAT} and \textbf{NT}]{
\includegraphics[width=\linewidth]{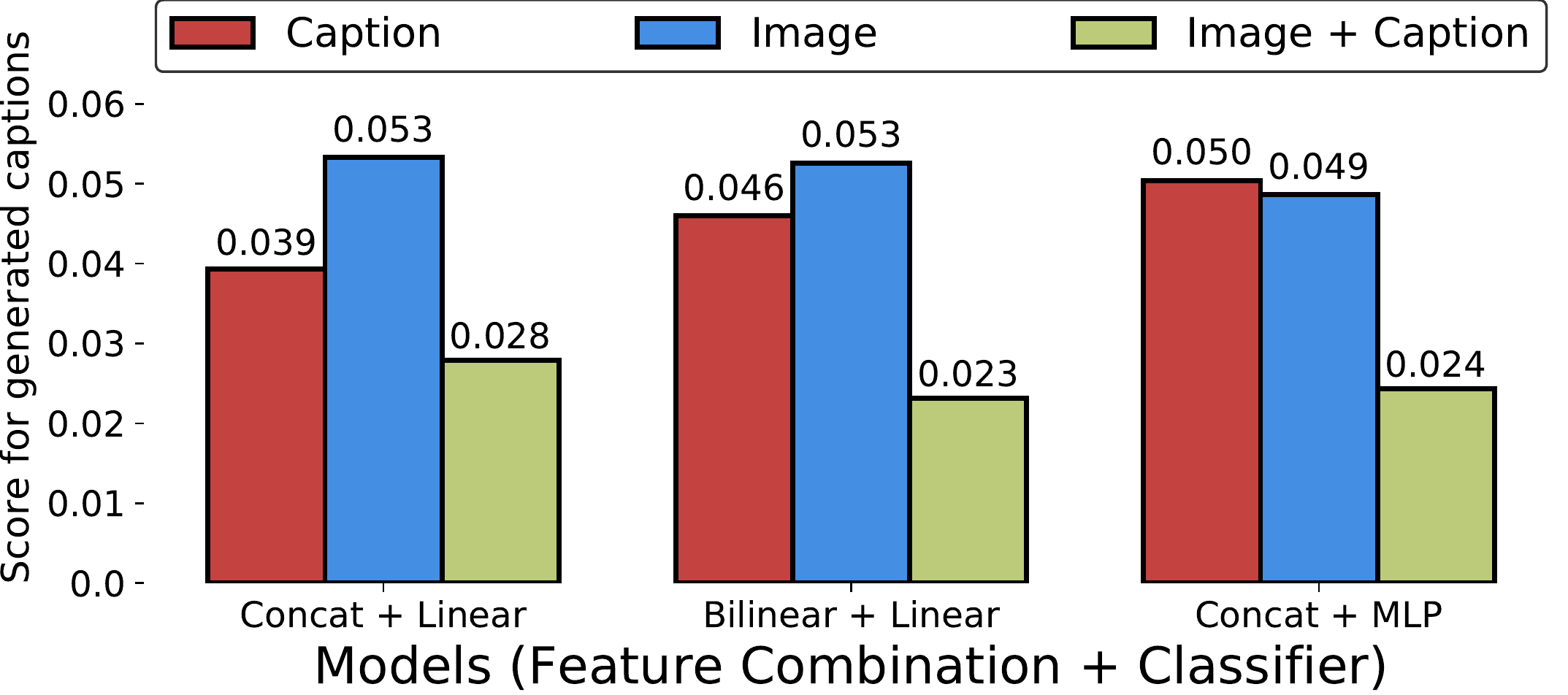}
   \label{fig:2_2}
 }
\caption{
This figure is same as Fig. 5 in the paper except all models are trained without data augmentation. 
} 
\label{fig:2}
\end{figure}

\section{The Choice of Hyper-parameters}
\label{sec:4.2}
Fig.~\ref{fig:1} compares capability performance of models with different LSTM layers and hidden feature sizes.
The proposed model is robust with respect to variant LSTM parameters.
Using models with higher capacity, \ie, more layers, higher dimensional hidden features, have no obvious benefit in terms of capability performance.
Considering the trade-off between performance gain and efficiency, we therefore use 1 LSTM layer and make the hidden feature of the LSTM to be 512 dimensional in our paper.

Fig.~\ref{fig:2} shows models trained without data augmentation.
Models trained with or without data augmentation are capable of learning to give higher scores to human captions than machine generated captions.
Interestingly, a critique trained without data augmentation can achieve even higher discrimination performance than models with data augmentation.
However, as shown in Sec. 4.3 and Fig. 6 in the paper, models trained without data augmentation are actually learning to perform a much simpler task, focusing only on discriminating human generated captions from the machine generated ones without considering the context (\ie, image and ground truth captions). 
Therefore, models that merely perform well in discrimination task might be easily gamed with pathological transformations.
Training with appropriate data augmentation and architecture (non-linearity) is essential to force critiques to pay attention to contexts.

\section{Caption Evaluation Examples}
\begin{figure*}[t]
\begin{center}
\includegraphics[width=\linewidth]{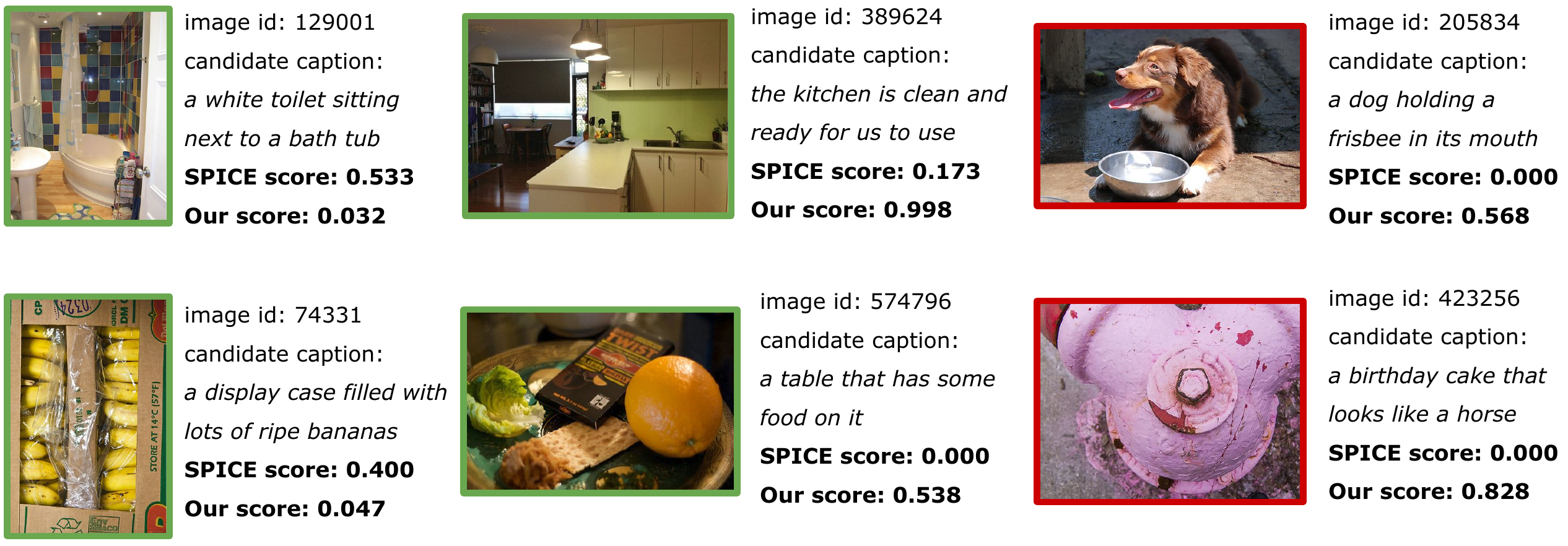}
\end{center}
   \caption{
   Exemplar candidate captions and their evaluation scores using our metric and SPICE on the COCO validation set. 
   }
\label{fig:example_captions}
\end{figure*}

Figure~\ref{fig:example_captions} provides examples captions of both success and failure cases.
Examples where our metric performs better than SPICE are marked with green bounding boxes, while examples where our metric is worse are marked with red ones. By utilizing the image as context, our metric is able to recognize some captions that are referring to wrong objects (left), and give high scores to captions that are semantically relevant to the image (center).
Typical failure cases of our metric are due to misleading visual information (right).

\section{System Level Human Correlation on COCO}
\label{sec:4.4}
In the original paper, we didn't compare to metrics M3, M4 and M5 because they were not used to rank image captioning models, but were intended for an ablation study to understand which aspects make captions good~\cite{cocochallenge}.
Since our metric was designed to evaluate the overall quality of an image caption, we only compared M1 and M2.
For better understanding of our metric from different perspectives, 
in Table~\ref{tab:1}, we calculate the Pearson's $\rho$ correlation between human judgements on all 5 metrics (M1-M5) used in 2015 COCO Captioning Challenge~\cite{cocochallenge}.

\begin{table*}[h]
\begin{center}
\begin{tabular}{ c|cc|cc|cc|cc|cc } 
\hline
 & \multicolumn{2}{c|}{\textbf{M1}} & \multicolumn{2}{c|}{\textbf{M2}} & \multicolumn{2}{c|}{\textbf{M3}} & \multicolumn{2}{c|}{\textbf{M4}} & \multicolumn{2}{c}{\textbf{M5}} \\ \cline{2-11}
 & $\rho$ & $p$-value & $\rho$ & $p$-value & $\rho$ & $p$-value & $\rho$ & $p$-value & $\rho$ & $p$-value \\
\hline
BLEU-1 & 0.124 & (0.687) & 0.135 & (0.660) & 0.549 & (0.052) & -0.517 & (0.070) & 0.241 & (0.428) \\
BLEU-2 & 0.037 & (0.903) & 0.048 & (0.877) & 0.483 & (0.094) & -0.572 & (0.041) & 0.162 & (0.598) \\
BLEU-3 & 0.004 & (0.990) & 0.016 & (0.959) & 0.471 & (0.105) & -0.588 & (0.035) & 0.143 & (0.641) \\
BLEU-4 & -0.019 & (0.951) & -0.005 & (0.987) & 0.459 & (0.114) & -0.577 & (0.039) & 0.139 & (0.650) \\
METEOR & 0.606 & (0.028) & 0.594 & (0.032) & 0.808 & (0.001) & 0.085 & (0.784) & 0.685 & (0.010) \\
ROUGE-L & 0.090 & (0.769) & 0.096 & (0.754) & 0.529 & (0.063) & -0.526 & (0.065) & 0.208 & (0.494) \\
CIDEr & 0.438 & (0.134) & 0.440 & (0.133) & 0.763 & (0.002) & -0.149 & (0.628) & 0.559 & (0.047) \\
SPICE & 0.759 & (0.003) & 0.750 & (0.003) & \textbf{0.871} & (0.000) & 0.250 & (0.411) & 0.809 & (0.001) \\
\textbf{Ours (no DA)} & \textbf{0.821} & (0.000) & \textbf{0.807} & (0.000) & 0.430 & (0.143) & \textbf{0.844} & (0.000) & 0.704 & (0.007) \\
\textbf{Ours} & \textbf{0.939} & (0.000) & \textbf{0.949} & (0.000) & 0.720 & (0.006) & \textbf{0.626} & (0.026) & \textbf{0.867} & (0.000) \\
\hline
\multicolumn{11}{l}{\textbf{M1}: Percentage of captions that are evaluated as better or equal to human caption.} \\
\hline
\multicolumn{11}{l}{\textbf{M2}: Percentage of captions that pass the Turing Test.} \\
\hline
\multicolumn{11}{l}{\textbf{M3 (Correctness)}: Average correctness of the captions on a scale 1-5 (incorrect - correct).} \\
\hline
\multicolumn{11}{l}{\textbf{M4 (Detailness)}: Average amount of detail of the captions on a scale 1-5 (lack of details - very detailed).} \\
\hline
\multicolumn{11}{l}{\textbf{M5 (Salience)}: Percentage of captions that are similar to human description.} \\
\hline
\end{tabular}
\end{center}
\caption{
Pearson's $\rho$ correlation between human judgements and evaluation metrics.
We use the 12 available entries to the 2015 MS-COCO captioning challenge that submitted results on the validation set.
``Ours (no DA)'' means our metric trained without data augmentation.
}
\label{tab:1}
\end{table*}

From the results, we can see that the human correlation of our proposed evaluation metrics surpasses all other metrics by large margins on M1, M2, M4 and M5.
On M3, our metric achieves comparable correlation scores with other commonly-used metrics.
It is worth noticing that all other metrics fail to capture the human correlation on the detailness of captions (M4), whereas our metric correlates reasonably well with humans on M4.

\section{How to Use the Proposed Metric in Practice}
We suggest the challenge organizer to first fix both the model architecture of the discriminator and all the hyper-parameters of the training process, then split the test set into two folds. 
This include fixing the number of training iterations for each submission.
For each submission, use the same setting to train the discriminator on one fold, and then use the trained metric to evaluate the other fold. Vise versa for the other split. After that, we get evaluation results on the full test set.
During training, the machine generated captions come from only the targeted submission, so that submission from one participants won't affect the score of the other participants.

\end{appendices}

\end{document}